\documentclass[twocolumn,10pt]{article}

\usepackage[T1]{fontenc}
\usepackage[utf8]{inputenc}
\usepackage{lmodern}
\usepackage{microtype}

\usepackage[margin=1in,columnsep=0.28in]{geometry}

\usepackage{amsmath}
\usepackage{amssymb,amsfonts}

\usepackage[numbers,sort]{natbib}

\usepackage{graphicx}
\usepackage{booktabs}
\usepackage{multirow}
\usepackage{float}
\usepackage[font=small,labelfont=bf]{caption}

\usepackage[hyphens]{url}
\usepackage{xcolor}
\definecolor{preprintblue}{RGB}{26,74,138}
\usepackage[unicode,colorlinks=true,linkcolor=preprintblue,%
            citecolor=preprintblue,urlcolor=preprintblue]{hyperref}

\usepackage[htt]{hyphenat}   %
\renewcommand{\_}{\textunderscore\allowbreak}
\DeclareUrlCommand\code{\urlstyle{tt}}

\newenvironment{acknowledgment}{\section*{Acknowledgment}}{}

\title{\bfseries Cultural Misalignment in Large Language Models:\\
Detection, Measurement, and Mitigation Through Targeted Fine-Tuning}

\author{%
  Antoni Czolgowski\\
  Masters in Data Science Program\\
  University of Colorado Boulder\\
  Boulder, CO 80309, USA\\
  \texttt{antoni.czolgowski@colorado.edu}
  \and
  Abel Iyasele\\
  Department of Information Science\\
  University of Colorado Boulder\\
  Boulder, CO 80309, USA\\
  \texttt{abel.iyasele@colorado.edu}%
}

\date{\today}

\begin{document}

\maketitle

\begingroup
\renewcommand{\thefootnote}{}%
\footnotetext{Extended version of a paper published in the proceedings of OSSConf 2026, Žilina, Slovakia.}%
\endgroup

\begin{abstract}
We evaluate three open-weight LLMs (Gemma3-12B from the USA, Bielik-11B-v3 from Poland, and Qwen3-4B from China) against World Values Survey Wave~7 data for 63 demographic personas across three countries, using normalized Wasserstein distance to quantify distributional misalignment. Contrary to expectations, no model favors its home country: the Chinese-built Qwen3-4B performs worst on its own Chinese population ($\hat{W}_1 = 0.436$, the highest misalignment in the entire model$\times$country matrix). Targeted LoRA fine-tuning on the five worst-case personas, requiring fewer than 1,200 training pairs and under 15 minutes on a single GPU, reduces bias by 16.8\% for Bielik-11B ($p_{\text{Bonf}} = 0.002$, $d = -4.4$) with all five targets improving. However, country-level decomposition reveals that fine-tuning redistributes rather than removes bias: Bielik's worst-case personas swap entirely from American to Chinese elderly, with zero overlap between pre- and post-correction sets. To our knowledge, this is the first study to target worst-case demographic personas with LoRA fine-tuning for cross-cultural bias mitigation.
\end{abstract}

\medskip


\section{Introduction}

Large language models are increasingly deployed in contexts that directly affect people's lives, from automated hiring systems and grant evaluations to public-facing chatbots and educational tools. Organizations across sectors embrace these systems as efficient and scalable, often without recognizing that the models carry systematic biases inherited from their training data and alignment procedures. When a language model consistently misrepresents the values or opinions of particular demographic groups, the consequences extend beyond technical inaccuracy: in the right context, such misrepresentation can translate into discriminatory outcomes for the very populations the technology is meant to serve.

A growing body of research has documented that LLMs do not treat all cultures equally. Studies have shown that models align better with Western, English-speaking populations and perform poorly on respondents from regions with limited representation in training corpora. However, the majority of this work focuses on commercial, closed-source models and limits its analysis to detecting bias without proposing concrete mitigation strategies. Furthermore, most existing approaches evaluate bias at the country or population level, overlooking the demographic subgroups within each country where misalignment may be most severe.

In this study, we address both gaps. We evaluate three open-weight models of different cultural provenance (Gemma3-12B from the United States, Bielik-11B-v3 from Poland, and Qwen3-4B from China) against human survey data from the World Values Survey Wave~7, constructing 63 demographic personas defined by country, sex, age group, and education level. Using the normalized Wasserstein distance as our bias metric, we quantify how well each model reproduces the response distributions of real human subpopulations. We then apply targeted LoRA fine-tuning to the worst-case personas and evaluate whether this intervention reduces bias without introducing unintended side effects on other demographic groups.

Our analysis is guided by three research questions: (1)~Do models exhibit systematic bias patterns related to their cultural origin? (2)~Does targeted fine-tuning on the most biased personas significantly reduce misalignment? (3)~Does such fine-tuning create collateral damage for non-targeted populations? The answers, as we show, challenge several intuitive assumptions about how cultural bias operates in language models.

The remainder of this paper is organized as follows. Section~2 reviews related work on cross-cultural bias, distributional alignment, and persona-based evaluation. Section~3 presents our methodology. Section~4 describes the experimental setup. Section~5 reports results for each research question. Section~6 discusses implications, limitations, and directions for future work.

\section{Related Work}

The question of whether language models carry cultural worldviews has evolved rapidly from a theoretical concern into an empirically grounded research program. This section traces that evolution across four interrelated themes: the emergence of cultural values as a lens for studying LLM behavior, the development of distributional metrics for measuring opinion alignment, the construction of demographic personas for survey simulation, and the application of fine-tuning methods for bias mitigation. Together, these threads reveal both the maturity of the measurement toolkit and the relative scarcity of work on targeted intervention, the gap our study addresses.

\subsection{Cultural Values and Language Models}

Large language models are not value-neutral instruments. Trained predominantly on English-language Internet corpora, they absorb the cultural assumptions, political orientations, and social norms embedded in those texts. Rozado~\cite{rozado2024politicalpreferencesllms} demonstrated that most modern conversational LLMs, including GPT-4 and models from the Llama family, exhibit left-leaning political preferences when evaluated on standardized political compass instruments, whereas their base (non-instruction-tuned) counterparts remain comparatively neutral. Crucially, these preferences crystallize during the fine-tuning stage, suggesting that post-training alignment procedures, rather than raw pretraining data alone, play a decisive role in shaping a model's expressed worldview.

The cultural dimension of this phenomenon extends well beyond the liberal-conservative axis. Tao et al.~\cite{tao2024cultural} conducted a disaggregated evaluation of five GPT-family models against nationally representative survey data from the Integrated Values Surveys and found that all models exhibited cultural values closely resembling those of English-speaking and Protestant European countries, regardless of the cultural context specified in the prompt. Their visualization on the Inglehart-Welzel World Cultural Map placed every tested model squarely in the self-expression and secular-rational quadrant, a region occupied by Scandinavian and Western European societies, but distant from the survival and traditional values characteristic of much of the Global South. Cultural prompting, the simple strategy of prepending a national identity to the query, improved alignment for 71--81\% of countries in later GPT versions, yet could not eliminate the underlying Western default entirely.

This phenomenon is not merely an artifact of training data volume; it reflects deliberate design choices. Reinforcement Learning with Human Feedback (RLHF), the dominant alignment paradigm, relies on preference annotations that inevitably encode the cultural backgrounds of the annotators involved~\cite{gallegos2024biasfairness}. When those annotators are predominantly drawn from Western, educated, industrialized, rich, and democratic (WEIRD) populations, the resulting reward signal systematically privileges one set of cultural assumptions over others. Barocas, Hardt, and Narayanan~\cite{barocas2023fairness} provide a broader theoretical treatment of how such design choices propagate through machine learning pipelines, arguing that fairness cannot be an afterthought bolted onto an already-trained system but must instead be considered at every stage of development.

The World Values Survey (WVS) has emerged as a particularly suitable instrument for quantifying this cultural misalignment. Spanning seven waves of data collection across more than 90 countries since 1981, the WVS provides nationally representative distributions of responses on topics ranging from religiosity and gender norms to democratic governance and economic attitudes. Its standardized multiple-choice format translates naturally into a distributional comparison framework: for any given question, one can compute the empirical response distribution from human survey data and compare it directly against the distribution of responses generated by a persona-prompted language model. Several foundational studies in LLM opinion alignment have adopted exactly this approach~\cite{Santurkar2023opinions, zhao2024worldvaluesbench, qu2024performance, liu2025alignment}, establishing the WVS as a cross-cultural benchmark that bridges social science methodology with computational evaluation.

An earlier study by one of the present authors~\cite{czolgowski2025bias} used WVS Wave~7 data to examine bias in the Gemma~2 2B model across eleven countries and five thematic domains. That work employed a regression-based methodology, constructing OLS models with dummy-coded demographic predictors and interpreting the magnitude of residuals between predicted survey responses and model outputs, and found that the model achieved substantially higher accuracy for respondents from developed Western countries (notably the Czech Republic and the United States) than for those from Libya, Nigeria, or India, with discrepancies exceeding 250\% on a logarithmic scale. While the regression approach provided interpretable coefficients for identifying which demographic groups were most poorly served, it treated each persona's prediction as a point estimate rather than a full distribution, limiting its ability to capture the heterogeneity of human opinion. The present study builds on that foundation by shifting to distribution-comparison metrics and adding a mitigation component absent from the earlier work.

\subsection{Bias Measurement and Distributional Analysis}

Early approaches to bias evaluation in language models focused primarily on detecting stereotypical associations: measuring, for instance, whether a model disproportionately associates certain occupations with particular genders or ethnicities~\cite{gallegos2024biasfairness}. While this line of work produced important benchmarks and raised public awareness, it addressed only the most overt manifestations of bias. A parallel and increasingly influential research program has shifted attention toward the subtler question of \emph{distributional alignment}: does the probability distribution over a model's responses to a subjective question match the distribution of responses observed in a human reference population?

The foundational contribution in this direction is the OpinionQA framework of Santurkar et al.~\cite{Santurkar2023opinions}, who compiled a dataset of 1,498 questions from Pew Research Center surveys spanning 15 topical domains and evaluated nine language models on their ability to reproduce the response distributions of various US demographic groups. Their key methodological innovation was the adoption of the 1-Wasserstein distance (also known as the Earth Mover's Distance) as the primary alignment metric. Unlike information-theoretic divergences such as Kullback-Leibler (KL) divergence or Jensen-Shannon divergence (JSD), the Wasserstein distance respects the ordinal structure of Likert-type response scales: shifting probability mass from ``Strongly Agree'' to ``Agree'' incurs a smaller penalty than shifting it to ``Strongly Disagree.'' This property is critical when working with ordered survey items, where adjacent response categories carry meaningful similarity that unordered divergence measures fail to capture. Santurkar et al.\ further introduced a normalized alignment score bounded between 0 and 1, computed as $A(D_M, D_H; Q) = \frac{1}{|Q|}\sum_{q \in Q}\left(1 - \frac{W_1(D_M(q), D_H(q))}{|N| - 1}\right)$, where $|N|$ is the number of response options and the denominator $|N|-1$ represents the maximum possible Wasserstein distance on that scale. This formulation has since become the standard metric in the field.

Subsequent work has refined both the measurement instruments and the methods for eliciting distributional information from models. Meister, Guestrin, and Hashimoto~\cite{meister2025benchmarking} conducted a systematic benchmark comparing three distribution expression methods: traditional model log-probabilities, sequential token sampling (generating a series of responses and computing the empirical frequency), and verbalized distributions (instructing the model to output a JSON object specifying the percentage of respondents expected to choose each option). Their analysis revealed that verbalized distributions consistently outperformed the other two methods, reducing the total variation distance to human reference distributions. However, the authors also identified a persistent ``knowledge-to-simulation gap'': models that could accurately \emph{describe} a group's likely opinion distribution in text often failed to \emph{sample} from that distribution when generating individual responses. This finding carries direct implications for studies like ours that rely on repeated sampling to construct empirical model distributions.

Extending the geographic scope beyond the United States, Liu, Kaneko, and Chu~\cite{liu2025alignment} created a comprehensive evaluation framework using all available WVS waves and tested seven state-of-the-art LLMs across 64 countries and eight prompt languages. Their country-level analysis revealed that models consistently achieved highest alignment with the United States, Chile, and Northern Ireland, while performing worst on Egypt, Myanmar, Libya, and Bangladesh, all of them countries characterized by non-Latin-script languages and limited representation in English-language training corpora. Notably, they found that changing the prompt language to match the target country improved alignment, although the effect was modest compared to the persistent baseline advantage enjoyed by Western nations. The WorldValuesBench dataset of Zhao et al.~\cite{zhao2024worldvaluesbench} provided a complementary resource by packaging WVS questions into a structured benchmark format with demographic conditioning, enabling standardized cross-model comparisons.

Several additional studies have enriched the empirical picture from different angles. Qu and Wang~\cite{qu2024performance} applied ChatGPT to simulate survey responses for six culturally diverse countries and found systematic overrepresentation of liberal and privileged viewpoints, with higher accuracy in English-speaking contexts. Lee et al.~\cite{lee2024warming} focused specifically on global warming opinions and documented significant demographic modulation of model accuracy. Ceron et al.~\cite{ceron2024political} examined the reliability and consistency of political worldviews expressed by LLMs, finding that model responses were sensitive to prompt rephrasing and answer choice ordering, a fragility that complicates alignment measurement. Sanders, Ulinich, and Schneier~\cite{sanders2023demonstrations} demonstrated the potential of AI-based political polling by comparing GPT-generated response distributions to those from the Cooperative Election Study, using normalized Earth Mover's Distance (the Wasserstein metric) to evaluate distributional similarity across ideological subgroups. Benkler et al.~\cite{benkler2023moralvalue} assessed moral value pluralism in LLMs and found limited capacity to represent non-Western moral frameworks. Sukiennik et al.~\cite{sukiennik2025evaluation} introduced the deviation ratio metric to evaluate cultural alignment despite a systematic tendency of LLMs to compress all cultures toward a moderate global average, and reported that the United States culture was most readily aligned by a wide margin, a finding they attributed in part to training data availability but also to deeper structural properties of how models encode cultural knowledge.

\subsection{Persona Construction for Survey Simulation}

A prerequisite for distributional bias measurement is the ability to condition a language model on a specific demographic profile, asking it to respond ``as'' a particular type of person. The construction and evaluation of such personas has become a research area in its own right, with implications for both the validity of LLM-based survey simulation and the design of fine-tuning interventions.

The choice of which demographic features to include in a persona prompt is consequential. Most studies converge on a core set of attributes (country, gender, age, and education), although some extend to income, political ideology, marital status, or employment~\cite{hwang2023aligning, kim2024aiaugmentedsurveys, qu2024performance}. The granularity of these features involves a tradeoff: finer categories improve the specificity of the simulated subgroup but reduce the number of human survey respondents available for comparison, potentially yielding unstable reference distributions. Deshpande et al.~\cite{deshpande2023toxicity} showed that even coarse persona assignments (e.g., instructing a model to respond ``as a man'' or ``as a woman'') can significantly alter the toxicity and content of generated text, confirming that LLMs are indeed sensitive to demographic framing.

The literature documents several distinct prompting strategies for demographic conditioning. Santurkar et al.~\cite{Santurkar2023opinions} and Zhao et al.~\cite{zhao2024worldvaluesbench} adopted a third-person framing (``Person X provided the following demographic information...''), which maintains analytical distance but may reduce the model's tendency to inhabit the persona. In contrast, studies following Argyle et al.'s influential ``silicon sampling'' paradigm use first-person role-play (``You are an interviewee. Based on your previous answers...''), which has been shown to elicit more differentiated responses~\cite{lee2024warming, qu2024performance}. A simpler direct assignment approach (``Please respond from the perspective of a 35-year-old Chinese male with higher education'') is used by Wright et al.~\cite{wright2025revealing} and Benkler et al.~\cite{benkler2023moralvalue}. Our study adopts a profile-plus-question format similar to that of Qu and Wang~\cite{qu2024performance}, in which demographic attributes are stated factually in a system prompt followed by the survey question, without elaborate backstory or role-play instructions. This neutral approach avoids predisposing the model toward culturally suggestive responses while still providing sufficient demographic conditioning.

A critical finding regarding the limits of persona prompting comes from Hu and Collier~\cite{hu2024personaeffect}, who quantified the proportion of variance in human annotations explained by persona variables (gender, age, education, etc.) across multiple NLP datasets. Their regression analysis revealed that persona variables accounted for less than 10\% of variance in most subjective tasks, a sobering result suggesting that demographic attributes alone are insufficient to capture the full heterogeneity of human opinion. This finding does not invalidate persona-based evaluation (the explained variance, while small, was statistically significant and consistent across tasks), but it does establish a realistic ceiling on how much distributional alignment can be achieved through demographic conditioning alone. Joshi et al.~\cite{joshi2024personas} explored this tension from a different angle, examining how persona assignments interact with model truthfulness and finding that certain persona configurations systematically altered a model's propensity for factual accuracy.

Temperature selection further modulates the diversity of model responses. The literature reflects a range of choices: Tao et al.~\cite{tao2024cultural} used temperature 0.0 for deterministic outputs, Qu and Wang~\cite{qu2024performance} used 0.2 for consistency, while Lee et al.~\cite{lee2024warming} and the majority of survey simulation studies adopt 0.7 as a standard compromise between reproducibility and response diversity. Sun et al.~\cite{sun2024siliconsampling} and Benkler et al.~\cite{benkler2023moralvalue} used 1.0 to maximize variance. The rationale for moderate temperatures in survey tasks is well established: when responses are constrained to predetermined ordinal options, temperature variations have limited effect on the content of individual outputs but meaningfully affect the shape of the resulting empirical distribution across repeated queries~\cite{qu2024performance}. Our study follows the 0.7 convention.

Several additional contributions have informed the broader understanding of how LLMs interact with survey methodology. Rosenbusch, Stevenson, and van der Maas~\cite{rosenbusch2023accurate} assessed the accuracy of GPT-3's hypotheses about social science phenomena and found reasonable but imperfect correspondence, suggesting that LLMs encode a useful but incomplete model of human social behavior. Smith-Vaniz et al.~\cite{smith2025moral} investigated political and demographic associations through the lens of Moral Foundations Theory, documenting systematic patterns in how models weight different moral dimensions based on assigned political identities. Dominguez-Olmedo, Hardt, and Mendler-D\"unner~\cite{dominguez2024survey} raised fundamental methodological concerns about treating LLM survey responses at face value, demonstrating that much of the apparent alignment between model and human distributions can be attributed to the model matching the majority response rather than genuinely representing the underlying heterogeneity, an observation that underscores the importance of distributional (rather than point-estimate) evaluation. Ma et al.~\cite{ma2024potentialchallenges} provided a comprehensive review of the potential and challenges of evaluating attitudes, opinions, and values in LLMs, proposing a four-stage evaluation taxonomy (input, model, output, evaluation) that has informed subsequent methodological work in the field, including the design of our own experimental pipeline.

\subsection{Fine-Tuning Methods for Bias Mitigation}

While the studies reviewed above have made substantial progress in \emph{detecting} and \emph{measuring} cultural bias, comparatively little work has addressed the question of \emph{mitigation}: how to systematically reduce the misalignment once it has been identified. The existing literature on bias mitigation in LLMs is extensive but predominantly focused on toxicity reduction and stereotype removal rather than on correcting distributional misalignment with specific cultural populations.

Gallegos et al.~\cite{gallegos2024biasfairness} provide a comprehensive taxonomy of bias mitigation strategies organized by the stage at which they intervene in the model pipeline. Pre-processing approaches modify training data through augmentation, filtering, reweighting, or instruction tuning. In-training methods alter the optimization process itself, whether through architecture modifications (e.g., adding debiasing adapter layers), loss function regularization that penalizes demographic disparities, selective parameter updating that freezes most weights to prevent catastrophic forgetting, or parameter pruning that removes neurons contributing to biased outputs. Intra-processing techniques modify inference-time behavior without retraining, such as adjusting the decoding strategy to reduce toxic token probabilities or applying modular debiasing networks that can be attached or detached as needed. Post-processing methods rewrite model outputs after generation. Each approach carries distinct advantages and limitations; our interest centers on in-training methods that modify model parameters, as these offer the most durable form of bias correction.

Among prompt-based mitigation strategies, cultural prompting, the approach of Tao et al.~\cite{tao2024cultural}, is notable for its simplicity and accessibility. By specifying a cultural identity in the prompt (e.g., ``Respond as someone from China''), it improved alignment for the majority of tested countries in GPT-4 and GPT-4o. However, the approach has fundamental limitations: it does not modify the model's internal representations, its effectiveness varies unpredictably across cultures and topics, and it requires the end user to specify the desired cultural frame at inference time. More robust alternatives require modifying model parameters.

Low-Rank Adaptation (LoRA), introduced by Hu et al.~\cite{hu2022lora}, has emerged as a particularly attractive fine-tuning strategy for bias mitigation. Rather than updating all model parameters, a process that is computationally expensive and risks catastrophic forgetting, LoRA freezes the pretrained weights and injects small, trainable low-rank matrices into each transformer layer. This approach reduces the number of trainable parameters by several orders of magnitude while preserving the model's general capabilities. Its parameter efficiency makes it feasible to train on consumer-grade hardware and to maintain multiple specialized adapters for different purposes, properties that are essential for a methodology intended to serve as a widely accessible benchmark.

Several studies have applied LoRA or related parameter-efficient methods to cultural alignment tasks. Li et al.~\cite{li2024culture} proposed CultureLLM, a framework that generates culture-specific training data by sampling seed questions from the WVS and applying semantic augmentation, then fine-tunes separate LoRA adapters for each target culture using supervised fine-tuning (SFT). Their approach demonstrated improved performance on downstream cultural tasks including offensive language detection and bias identification, establishing that relatively small amounts of culturally grounded training data can measurably shift a model's cultural orientation. Jinnai~\cite{jinnai2024morality} investigated cross-cultural alignment through Direct Preference Optimization (DPO), an alternative to RLHF that optimizes directly on preference pairs without training a separate reward model. A key finding of that work was that alignment with cultural commonsense morality proved more important than alignment with the target language: a model fine-tuned on English-language data reflecting Japanese moral norms outperformed one fine-tuned on Japanese-language data reflecting Western norms. This result suggests that the \emph{cultural content} of fine-tuning data matters more than its \emph{linguistic form}, with implications for our own study's use of English-language WVS data to fine-tune models with diverse cultural provenances.

Haller, Aynetdinov, and Akbik~\cite{haller2024opiniongpt} took a different approach with OpinionGPT, training eleven separate LoRA adapters on a Llama~2 13B base model, each adapter corresponding to a distinct demographic bias (geographic, political, gender, or age). Their Mixture-of-Experts design allows users to select which perspective they wish the model to adopt, making biases explicit rather than suppressing them. The authors noted that dedicated LoRA adapters combined with a larger base model captured demographic-specific biases more precisely than full fine-tuning of a smaller model, an observation relevant to our finding, discussed later, that model scale influences receptiveness to LoRA-based alignment corrections. Indeed, recent evidence suggests a positive correlation between model size and both baseline cultural alignment and fine-tuning responsiveness: Liu et al.~\cite{liu2025alignment} found that alignment scores increased monotonically with model size across the Qwen2.5 family (0.5B to 14B parameters), and Bai et al.~\cite{bai2025biased} reported that larger models retained richer implicit associations in their representations despite explicit debiasing, suggesting greater representational capacity that fine-tuning can leverage. We revisit the relationship between model scale and fine-tuning efficacy in the context of our experimental results.

Evidence from adjacent domains further supports the viability of fine-tuning for bias reduction. Huang et al.~\cite{huang2024collectiveAI} demonstrated that Constitutional AI, a form of RLHF guided by publicly sourced principles, reduced bias across nine social dimensions on the BBQ benchmark while maintaining general capabilities (MMLU scores remained within 0.1\% of the unmodified model). Li et al.~\cite{li2024psychological} evaluated the psychological safety of fine-tuned LLMs using personality inventories and found that DPO-based fine-tuning could effectively reduce dark personality traits, confirming that targeted optimization can reshape specific behavioral patterns without global degradation. However, An et al.~\cite{an2025gender} sounded an important cautionary note: their analysis of gender and racial bias in resume evaluation revealed that newer models attempting to correct historical biases had ``overcorrected,'' creating preferential treatment for previously disadvantaged groups and thereby generating new forms of inequity. This overcorrection problem, where fixing bias in one direction creates bias in another, motivates the explicit side-effect evaluation component (RQ3) of our study design.

The measurement modeling framework of Jacobs and Wallach~\cite{jacobs2021} provides conceptual grounding for interpreting these mitigation efforts. Their central argument, that ``bias'' is an essentially contested construct whose operationalization depends on the theoretical framework adopted, reminds us that distributional alignment as measured by the Wasserstein distance captures one specific aspect of cultural bias (the distance between a model's opinion distribution and a human reference distribution) but does not exhaust the phenomenon. Different metrics, different survey instruments, and different reference populations would yield different conclusions about which models are ``biased'' and in what ways. We adopt the Wasserstein framework not because it is uniquely correct, but because it is well-suited to ordinal survey data, mathematically well-founded in optimal transport theory, and directly comparable to the existing body of alignment research.

\medskip

\noindent\textbf{Research gap.} Despite the substantial progress reviewed above, no prior work combines all four elements necessary for a complete bias detection-and-mitigation pipeline: (a)~a distributional metric that respects ordinal structure (the normalized Wasserstein $W_1$ distance), (b)~systematic identification of worst-case demographic personas through bootstrap confidence interval analysis, (c)~targeted LoRA fine-tuning applied specifically to those worst-case personas, and (d)~rigorous side-effect evaluation decomposed by country to ensure that correcting bias for one population does not degrade alignment for others. Furthermore, existing cross-cultural studies have relied exclusively on models from a single geopolitical origin (predominantly US-based), leaving open the question of whether a model's cultural provenance systematically shapes its bias profile. Our study addresses these gaps by evaluating three models originating from three distinct geopolitical contexts, Gemma3-12B (Google, USA), Bielik-11B-v3 (SpeakLeash, Poland), and Qwen3-4B (Alibaba, China), against WVS Wave~7 data from the United States, Slovakia (serving as a Central European proxy for Poland, given the absence of Polish data in Wave~7; both nations share extensive historical ties as Visegrad Group members and occupy adjacent positions on cross-cultural value maps), and China. Slovakia's suitability as a proxy is supported by the shared post-socialist transition experience, comparable religiosity profiles, and geographic proximity that produce closely aligned WVS response distributions on multiple value dimensions. To our knowledge, this is the first study to target worst-case demographic personas with LoRA fine-tuning for cross-cultural bias mitigation, proposing a reproducible benchmark methodology designed for model creators conducting pre-deployment bias assessment.

\subsection{Research Questions}

The gaps identified above motivate three research questions that together form a detection--mitigation--evaluation pipeline:

\begin{enumerate}
\item[\textbf{RQ1}] \textbf{(Cultural Origin Bias):} Do language models systematically favor demographic personas aligned with their country of origin or training provenance?
\item[\textbf{RQ2}] \textbf{(Bias Mitigation Efficacy):} Does targeted LoRA fine-tuning on worst-case demographic personas significantly reduce bias, measured by Wasserstein distance, without overfitting to the training distribution?
\item[\textbf{RQ3}] \textbf{(Mitigation Side Effects):} Does fine-tuning to improve worst-case personas introduce negative side effects on model performance for other demographic groups?
\end{enumerate}

RQ1 tests whether training data composition shapes a model's cultural orientation by comparing three models of distinct geopolitical origin (American, Polish, and Chinese) against personas from the United States, Slovakia, and China. If cultural provenance matters, models should exhibit lower distributional distance for culturally proximate populations.

RQ2 addresses the mitigation gap identified in the literature. Our strategy rests on the assumption that misalignment stems partly from underrepresentation of specific demographic perspectives. We construct fine-tuning data from \emph{other} WVS questions answered by the same worst-case personas, deliberately excluding the evaluation question to prevent data leakage, so that any improvement on the held-out question reflects genuine cultural learning rather than memorization.

RQ3 responds to the overcorrection problem documented by An et al.~\cite{an2025gender}: fixing bias for one group may create new disparities for others. We evaluate side effects not only in aggregate but decomposed by country, since redistribution effects can remain invisible in summary statistics.

\section{Methodology}

This section describes the five components of our experimental methodology: the selection of a suitable survey question from the World Values Survey, the construction of demographically conditioned personas, the distributional metric used to quantify bias, the selection of models from three geopolitical origins, and the fine-tuning procedure applied to the worst-performing personas.

\subsection{Data Source and Question Selection}
 
We draw our human reference data from the World Values Survey (WVS) Wave~7, a cross-national research program that collected standardized survey responses from 93,728 respondents across 64 countries between 2017 and 2022. The WVS has been widely adopted as a benchmark for evaluating cultural alignment in language models~\cite{Santurkar2023opinions, zhao2024worldvaluesbench, tao2024cultural, liu2025alignment}, owing to its nationally representative sampling, standardized multiple-choice format, and broad thematic coverage spanning religiosity, political attitudes, social norms, and economic values.
 
From the WVS instrument, we select question Q164: ``How important is God in your life?'' rated on a scale from 1 (not at all important) to 10 (very important). This question was chosen for three reasons. First, it exhibits maximum cross-cultural variance among the candidate items we considered: China's mean response is 2.8 (strongly secular, with 52\% of respondents answering ``1''), Slovakia's is 6.4 (moderately religious), and the United States' is 6.7 (highly religious, with 39\% answering ``10''). This range spans nearly the entire scale and ensures that models must produce genuinely different distributions to align with each country's population. Second, the 1--10 ordinal scale is well suited to the Wasserstein distance metric (Section~\ref{sec:wasserstein}), which exploits the ordered structure of response categories. Third, Q164 has been used in prior LLM evaluation studies~\cite{tao2024cultural, benkler2023moralvalue}, enabling direct comparison with existing findings.
 
We considered two alternative questions, Q182 (justifiability of homosexuality) and Q184 (justifiability of abortion), which also exhibit strong cross-cultural variance. However, both carry a higher risk of triggering model safety filters or refusal behaviors, particularly for models with aggressive RLHF alignment. Q164, while culturally sensitive in its own right, proved empirically unproblematic: all three models in our study produced valid integer responses with a 0\% refusal rate across 31,500 total queries.
 
Data cleaning followed standard WVS protocols: responses coded as ``Don't know,'' ``No answer,'' or ``Refused'' were removed prior to computing empirical distributions. Table~\ref{tab:country_profiles} summarizes the resulting country profiles.
 
\begin{table*}[tp]
    \centering
    \small
    \caption{WVS Wave~7 country profiles for Q164 (``Importance of God''). Slovakia serves as the Central European proxy for Poland, which is absent from Wave~7.}
    \label{tab:country_profiles}
    \begin{tabular}{llcccc}
        \toprule
        Country & Code & Mean & Std Dev & Character & $N_{\text{WVS}}$ \\
        \midrule
        China & CHN & 2.8 & ${\sim}2.5$ & Strongly secular & ${\sim}3{,}000$ \\
        Slovakia & SVK & 6.4 & ${\sim}3.3$ & Moderately religious & ${\sim}1{,}200$ \\
        United States & USA & 6.7 & ${\sim}3.4$ & Highly religious & ${\sim}2{,}500$ \\
        \bottomrule
    \end{tabular}
\end{table*}

A critical design consideration concerns the choice of countries. Our study evaluates three models originating from three distinct geopolitical regions: the United States (Gemma3-12B, Google), Poland (Bielik-11B-v3, SpeakLeash), and China (Qwen3-4B, Alibaba). Testing whether models favor their country of origin (RQ1) therefore requires human reference populations from corresponding cultural contexts. However, Poland does not appear in WVS Wave~7. We therefore use Slovakia as a culturally proximate substitute, justified by several considerations: both nations are post-socialist Central European states that share membership in the Visegrad Group (V4), occupy adjacent positions on the Inglehart-Welzel World Cultural Map, and exhibit comparable religiosity profiles shaped by their shared Catholic heritage and parallel transition experiences since 1989. Throughout this paper, all references to WVS data for the Central European cultural region refer to Slovak respondents; we make no claim that Slovak and Polish populations are identical, only that Slovakia represents the closest available proxy within the WVS Wave~7 data.

\subsection{Persona Construction}
\label{sec:personas}
 
To evaluate model bias at the demographic subgroup level, we construct personas by crossing four demographic features drawn from the WVS respondent metadata: country (China, Slovakia, United States), sex (male, female), age group (18--29, 30--49, 50--64, 65+), and education level (lower, medium, higher). This yields a theoretical design space of $3 \times 2 \times 4 \times 3 = 72$ unique personas.
 
We initially considered including a fifth feature, subjective social class (upper, upper middle, lower middle, working, lower), which would have expanded the space to 360 combinations. However, preliminary analysis of the WVS data revealed that upper-class cells were severely underrepresented across all three countries, with many combinations containing fewer than five respondents. We therefore excluded social class from the persona design to ensure stable reference distributions.
 
Education levels follow the International Standard Classification of Education (ISCED): lower education encompasses ISCED levels 0--2 (primary and lower secondary), medium encompasses ISCED 3--4 (upper secondary and post-secondary non-tertiary), and higher encompasses ISCED 5--8 (tertiary education through doctoral level).
 
Not all 72 theoretical personas have sufficient human respondent data for reliable distributional comparison. We impose a minimum threshold of $N \geq 10$ WVS respondents per persona, below which the empirical probability mass function (PMF) over the 10-point scale becomes too sparse for meaningful Wasserstein distance computation. Nine of the 72 combinations fell below this threshold and were excluded, yielding \textbf{63 valid personas}: 23 Chinese, 23 Slovak, and 17 American. The disproportionate exclusion of American personas reflects lower WVS sample sizes in certain age--education intersections for the US subsample.
 
Table~\ref{tab:persona_design} summarizes the persona construction design. Each persona is identified by a compound label encoding all four features (e.g., \texttt{USA\_Male\_30-49\_Medium}).
 
\begin{table*}[tp]
    \centering
    \caption{Persona construction design. The four demographic dimensions are fully crossed, subject to a minimum sample size threshold of $N \geq 10$ WVS respondents per cell.}
    \label{tab:persona_design}
    \small
    \begin{tabular}{l p{9.5cm} l}
        \toprule
        Dimension & Categories & Source \\
        \midrule
        Country & China (CHN), Slovakia (SVK), USA & WVS B\_COUNTRY \\
        Sex & Male, Female & WVS Q260 \\
        Age group & 18--29, 30--49, 50--64, 65+ & WVS Q262 (binned) \\
        Education & Lower (ISCED 0--2), Medium (3--4), Higher (5--8) & WVS Q275R \\
        \midrule
        \multicolumn{3}{l}{Theoretical combinations: $3 \times 2 \times 4 \times 3 = 72$} \\
        \multicolumn{3}{l}{Valid personas ($N \geq 10$): 63 (23 CHN, 23 SVK, 17 USA)} \\
        \bottomrule
    \end{tabular}
\end{table*}
 
Each persona is queried using a structured prompt that specifies the demographic profile in a system message followed by the survey question. Figure~\ref{fig:prompt_template} in Appendix~A illustrates the complete prompting pipeline, showing how demographic attributes from the WVS are translated into a persona card and then into the structured prompt format. The prompt design follows the profile-plus-question paradigm described by Qu and Wang~\cite{qu2024performance}, in which demographic attributes are stated factually without elaborate role-play or backstory, and the model is instructed to respond with only a single integer. This neutral approach avoids predisposing the model toward culturally suggestive responses while providing sufficient demographic conditioning~\cite{deshpande2023toxicity}. We note that Hu and Collier~\cite{hu2024personaeffect} found that persona variables account for less than 10\% of variance in most subjective NLP tasks, establishing a realistic ceiling on achievable alignment through demographic conditioning alone.

\subsection{Wasserstein Distance Metric}
\label{sec:wasserstein}
 
To quantify the distributional distance between human survey responses and model outputs, we adopt the 1-Wasserstein distance ($W_1$), also known as the Earth Mover's Distance (EMD). This metric, grounded in optimal transport theory~\cite{villani2008optimal}, measures the minimum cost of transforming one probability distribution into another, where cost is defined as the amount of probability mass moved multiplied by the distance it is transported.
 
For discrete ordered distributions on the support $\{1, 2, \ldots, 10\}$, the $W_1$ distance admits a closed-form expression in terms of cumulative distribution functions (CDFs):
\begin{equation}
    W_1(P, Q) = \sum_{k=1}^{9} \left| F_P(k) - F_Q(k) \right|
    \label{eq:w1}
\end{equation}
where $F_P$ and $F_Q$ are the CDFs of the human and model response distributions, respectively. The summation runs from $k=1$ to $k=9$ because the CDF at $k=10$ is identically 1 for both distributions.
 
The choice of $W_1$ over information-theoretic alternatives such as Kullback-Leibler (KL) divergence or Jensen-Shannon divergence (JSD) is motivated by the ordinal structure of our response scale. As Santurkar et al.~\cite{Santurkar2023opinions} argued in their foundational OpinionQA framework, the Wasserstein distance respects the ordering of response categories: shifting probability mass from ``Strongly Agree'' to ``Agree'' incurs a smaller penalty than shifting it to ``Strongly Disagree.'' KL and JSD treat all category mismatches equally, which is inappropriate for Likert-type items where adjacent categories carry meaningful similarity. Meister et al.~\cite{meister2025benchmarking} confirmed the superiority of distributional metrics over point estimates for opinion alignment evaluation.
 
Following Santurkar et al.~\cite{Santurkar2023opinions}, we normalize $W_1$ by dividing by 9, the maximum possible distance on a 1--10 scale (achieved when one distribution places all mass at 1 and the other at 10):
\begin{equation}
    \hat{W}_1 = \frac{W_1}{9}
    \label{eq:w1_norm}
\end{equation}
This yields a normalized score $\hat{W}_1 \in [0, 1]$, where 0 indicates perfect distributional alignment and 1 indicates maximum divergence. All $W_1$ values reported in this paper are normalized unless otherwise noted. Higher $\hat{W}_1$ indicates greater bias (i.e., worse alignment with the human reference population).
 
\begin{figure*}[tp]
    \centering
    \includegraphics[width=0.62\textwidth]{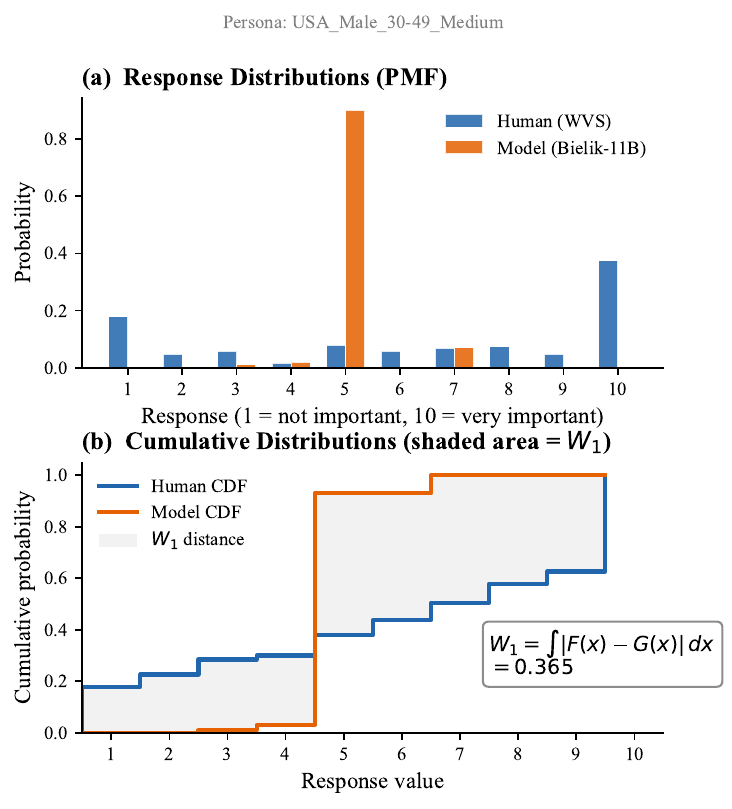}
    \caption{Illustration of the Wasserstein $W_1$ distance for the persona \texttt{USA\_Male\_30-49\_Medium} (Bielik-11B). (a)~Response distributions (PMFs): the human distribution (blue) is broadly dispersed with a mode at 10, while the model (orange) concentrates at 5. (b)~Cumulative distributions: the shaded area between the CDFs equals the $W_1$ distance ($\hat{W}_1 = 0.365$).}
    \label{fig:w1_explanation}
\end{figure*}

Figure~\ref{fig:w1_explanation} illustrates the metric using a concrete example from our data: the persona \texttt{USA\_Male\_30-49\_Medium} evaluated against Bielik-11B. Panel~(a) shows the response probability mass functions (PMFs), revealing a stark contrast: the human distribution is broadly dispersed across the scale with a strong mode at 10 (37\% of respondents), reflecting the religiosity typical of American middle-aged males, while the model concentrates 90\% of its probability mass at the value 5, producing a ``fence-sitting'' response that fails to capture the heterogeneity of human opinion. Panel~(b) overlays the corresponding CDFs; the shaded area between the two step functions equals $W_1 = 0.365$ (normalized), quantifying the total distributional misalignment.
 
\paragraph{Bootstrap confidence intervals.} Human survey responses are subject to sampling variability: a persona with $N = 15$ WVS respondents yields a less stable empirical PMF than one with $N = 350$. To account for this, we construct 95\% confidence intervals for $\hat{W}_1$ using a nonparametric bootstrap procedure. For each persona, we resample the $N_{\text{WVS}}$ human responses with replacement 1{,}000 times, recompute the empirical PMF and the resulting $\hat{W}_1$ against the (fixed) model distribution at each iteration, and take the 2.5th and 97.5th percentiles of the bootstrap distribution as the CI bounds. The model side uses the raw empirical distribution from 100 queries and is not bootstrapped, as it represents a fixed experimental quantity rather than a sample from a larger population. Figure~\ref{fig:ci_vs_n} in Appendix~A confirms the expected relationship: CI width decreases monotonically with human sample size, from approximately 0.30 for the smallest personas ($N \approx 10$) to below 0.05 for the largest ($N > 300$).
 
\paragraph{Worst-case persona selection.} For the fine-tuning intervention (Section~\ref{sec:finetuning}), we must identify the personas that exhibit the most severe bias. Rather than ranking by point estimates of $\hat{W}_1$, which would be unreliable for small-$N$ personas with wide confidence intervals, we select the top~5 worst-case personas per model based on the \emph{lower bound} of the 95\% bootstrap CI. This conservative criterion ensures that a persona is flagged as worst-case only when we are confident that its true $\hat{W}_1$ is high, not merely when sampling noise produces a single large estimate. This approach trades sensitivity for specificity: it may miss some genuinely problematic personas with small samples, but the personas it does identify are robustly misaligned.

\subsection{Model Selection}
\label{sec:models}
 
Our study evaluates three open-weight large language models, each originating from a distinct geopolitical and cultural context. This selection is designed to test RQ1 (whether a model's cultural provenance shapes its bias profile) by comparing models whose training data, organizational culture, and alignment procedures reflect the norms of different societies.

\begin{table*}[tp]
    \centering
    \small
    \caption{Model specifications. All models are open-weight and locally deployable on a single NVIDIA A100 GPU.}
    \label{tab:models}
    \begin{tabular}{llrlll}
        \toprule
        Model & Origin & Params & Arch. & Quant. & Role \\
        \midrule
        Gemma3-12B & USA (Google) & 12B & Gemma & None (Ollama) & Baseline \\
        Bielik-11B-v3 & Poland (SpeakLeash) & 11B & Mistral & 4-bit BnB & Baseline + LoRA \\
        Qwen3-4B & China (Alibaba) & 4B & Qwen & 4-bit BnB & Baseline + LoRA \\
        \bottomrule
    \end{tabular}
\end{table*}
 
\textbf{Gemma3-12B} (Google, USA) is an instruction-tuned model from Google's Gemma family, representing the dominant US-centric paradigm in LLM development. With 12 billion parameters, it is the largest model in our study. We deploy it without quantization via the Ollama inference framework, which provides a containerized serving interface. Gemma3 serves as a baseline-only model: because Ollama does not expose the underlying HuggingFace model weights, we cannot apply LoRA fine-tuning through the standard PEFT pipeline.
 
\textbf{Bielik-11B-v3} (SpeakLeash, Poland) is a Mistral-architecture model developed by the SpeakLeash consortium, a Polish open-source initiative focused on creating language technology for the Polish language. At 11 billion parameters, it is the closest in scale to Gemma3. We deploy it via HuggingFace Transformers with 4-bit quantization (BitsAndBytes), which reduces memory requirements to fit within a single GPU while preserving response quality for survey-type tasks. Bielik is included both in the baseline evaluation and as a LoRA fine-tuning target.
 
\textbf{Qwen3-4B-Instruct-2507} (Alibaba, China) is the smallest model in our study at 4 billion parameters. Developed by Alibaba's Qwen team, it represents the rapidly growing ecosystem of Chinese-origin LLMs. Like Bielik, it is deployed via HuggingFace Transformers with 4-bit quantization and serves as both a baseline and LoRA fine-tuning target.
 
The selection of these specific models was guided by three criteria: (a)~distinct cultural provenance, ensuring that RQ1 can be meaningfully tested; (b)~availability for local deployment without API access, which is essential for reproducibility and for running experiments on an HPC cluster; and (c)~feasibility on a single-GPU setup, reflecting our design choice to demonstrate a methodology that is accessible and reproducible by researchers without large-scale compute infrastructure. We deliberately chose LoRA as the fine-tuning method, rather than full fine-tuning or DPO, because parameter-efficient methods are compatible with single-GPU constraints and represent the most practically accessible option for the model creators who constitute our intended audience.
 
The parameter count disparity between models (4B vs.\ 11B vs.\ 12B) is worth noting. While this introduces a confound in direct model-to-model comparisons, it also creates an opportunity to examine whether model scale affects receptiveness to LoRA fine-tuning, a relationship that has been suggested in prior work. Liu et al.~\cite{liu2025alignment} reported monotonically increasing alignment scores with model size across the Qwen2.5 family, and Haller et al.~\cite{haller2024opiniongpt} observed that larger base models captured demographic-specific biases more precisely with LoRA adapters. We return to this question in the Discussion.

\subsection{Fine-Tuning Procedure}
\label{sec:finetuning}
 
Our fine-tuning intervention targets the five worst-case personas per model, identified through the bootstrap-based selection procedure described in Section~\ref{sec:wasserstein}. The goal is to shift each model's response distribution closer to the human reference for those specific demographic groups, without degrading performance on other personas (RQ3).
 
\paragraph{Training data construction.} A critical design decision concerns the source of fine-tuning data. If we were to train on Q164 responses directly, any improvement in $\hat{W}_1$ on Q164 could reflect memorization rather than genuine cultural learning. We therefore construct training data exclusively from \emph{non-religious} WVS Wave~7 questions, deliberately excluding Q164 and all other religiosity-related items from the training set. This creates a transfer learning setup: the model learns the response patterns of the target personas from their answers to questions about politics, economics, social norms, and family values, and we evaluate whether this cultural knowledge transfers to the held-out religiosity question.
 
For each worst-case persona, we extract all available non-religious WVS questions for which the persona's demographic subgroup has sufficient data ($N \geq 10$). Each training example is formatted as a supervised fine-tuning (SFT) pair: the input is a persona-conditioned prompt (following the same template used for evaluation; see Figure~\ref{fig:prompt_template} in Appendix~A), and the target output is the human modal response for that question and demographic group. This produces persona-specific training data that reflects the actual opinion distribution of the target population across a broad range of topics.
 
The resulting training sets differ in size between models because the worst-case personas come from different countries with different WVS question coverage:
\begin{itemize}
    \item \textbf{Bielik:} 5 worst-case personas from the United States, yielding 1{,}180 training pairs (5 personas $\times$ 236 non-religious questions).
    \item \textbf{Qwen:} 5 worst-case personas from China, yielding 1{,}003 training pairs (5 personas $\times$ approximately 201 non-religious questions per persona). The lower count reflects fewer available WVS questions for the Chinese subsample.
\end{itemize}
 
\paragraph{LoRA configuration.} We use Low-Rank Adaptation (LoRA)~\cite{hu2022lora} to fine-tune each model. LoRA freezes all pretrained weights and injects small, trainable low-rank matrices into selected transformer layers, reducing the number of trainable parameters by several orders of magnitude while preserving general model capabilities. Our configuration targets the query and value projection matrices (\texttt{q\_proj} and \texttt{v\_proj}) in all attention layers, with rank $r = 16$, scaling factor $\alpha = 32$ (yielding an effective learning rate multiplier of $\alpha / r = 2$), and dropout rate 0.05. Training runs for 3 epochs with a learning rate of $2 \times 10^{-4}$ and batch size of 4.
 
Table~\ref{tab:lora_config} summarizes the LoRA configuration and training statistics for both models.
 
\begin{table}[htbp]
    \centering
    \caption{LoRA fine-tuning configuration and training statistics. Gemma3-12B is excluded because its Ollama deployment does not expose model weights for fine-tuning.}
    \label{tab:lora_config}
    \small
    \setlength{\tabcolsep}{3.5pt}
    \begin{tabular}{lcc}
        \toprule
        & Bielik-11B & Qwen3-4B \\
        \midrule
        Target personas & 5 (USA) & 5 (CHN) \\
        Training pairs & 1{,}180 & 1{,}003 \\
        LoRA rank ($r$) & 16 & 16 \\
        LoRA alpha ($\alpha$) & 32 & 32 \\
        LoRA dropout & 0.05 & 0.05 \\
        Target modules & q\_proj, v\_proj & q\_proj, v\_proj \\
        Epochs & 3 & 3 \\
        Learning rate & $2 \times 10^{-4}$ & $2 \times 10^{-4}$ \\
        Training time & 13.6 min & 4.9 min \\
        Loss (start $\to$ end) & $2.53 \to 0.39$ & $1.13 \to 0.36$ \\
        \bottomrule
    \end{tabular}
\end{table}
 
The choice of LoRA over alternative fine-tuning methods, such as full fine-tuning or Direct Preference Optimization (DPO)~\cite{jinnai2024morality}, is a deliberate methodological decision. Our aim is to propose a benchmark methodology that is accessible to the broadest possible range of model creators, including those without access to multi-GPU clusters. LoRA's parameter efficiency (training completes in under 15 minutes on a single A100) and its compatibility with quantized models make it the most practically viable option for the pre-deployment bias assessment pipeline we envision. We note that CultureLLM~\cite{li2024culture} demonstrated effective cultural alignment using LoRA with similarly modest training sets, and OpinionGPT~\cite{haller2024opiniongpt} confirmed that LoRA adapters can capture demographic-specific opinion patterns in models of comparable scale.
 
After fine-tuning, we re-evaluate each model on all 63 personas using the same query protocol as the baseline evaluation (Section~\ref{sec:query_protocol}), generating 100 new responses per persona. This produces a complete post-LoRA evaluation dataset that enables both targeted comparison (RQ2: did the 5 worst-case personas improve?) and side-effect analysis (RQ3: did the remaining 58 personas change?).

\section{Experimental Setup}

\subsection{Computational Infrastructure}
 
All experiments were conducted on the Alpine high-performance computing cluster at the University of Colorado Boulder, operated by CU Research Computing (CURC). The cluster provides NVIDIA A100 80\,GB GPUs on the \texttt{aa100} partition, with a maximum job duration of 24 hours. Our allocation provides 350{,}000 Service Units (SUs), where GPU usage is billed at 108 SU per hour.
 
The computational infrastructure uses two distinct model serving pipelines depending on the model's deployment requirements:
\begin{itemize}
    \item \textbf{Ollama} serves Gemma3-12B as a containerized inference endpoint. Because Alpine assigns dynamic ports per session (rather than the default 11434), the \code{OLLAMA_HOST} environment variable is read at runtime to ensure correct routing. Ollama's quantization-free serving provides native model precision but does not expose HuggingFace-compatible model weights, precluding LoRA fine-tuning through the PEFT library.
    \item \textbf{HuggingFace Transformers} serves Bielik-11B-v3 and Qwen3-4B-Instruct-2507 from locally stored weights. Both models are loaded with 4-bit quantization via BitsAndBytes (\code{load_in_4bit=True}, \code{bnb_4bit_compute_dtype=bfloat16}), reducing GPU memory requirements to approximately 6--8\,GB per model and enabling all inference and fine-tuning on a single A100 GPU.
\end{itemize}
 
 The software environment consists of a Conda environment running Python 3.10 with PyTorch, Transformers, Accelerate, PEFT, and BitsAndBytes. Statistical analysis was performed in R~4.5.2 using the tidyverse ecosystem. All code and data are publicly available online.\footnote{\url{https://github.com/AntoniCzolgowski/llm-cultural-bias}}

\subsection{Query Protocol}
\label{sec:query_protocol}
 
Each of the 63 valid personas is queried 100 times per model, producing an empirical response distribution over the 1--10 scale. The baseline evaluation generates $63 \times 100 \times 3 = 18{,}900$ queries across the three models; the post-LoRA evaluation generates an additional $63 \times 100 \times 2 = 12{,}600$ queries for Bielik and Qwen, for a total of \textbf{31{,}500 queries} across all experimental phases.
 
All queries use the prompt template illustrated in Figure~\ref{fig:prompt_template} (Appendix~A), with generation parameters set to temperature 0.7, top-$k$ = 50, and maximum new tokens = 10. The temperature of 0.7 follows the convention established in the survey simulation literature~\cite{lee2024warming, qu2024performance} as a compromise between response diversity (necessary to produce meaningful empirical distributions) and output stability (avoiding incoherent responses). The maximum token limit of 10 constrains the model to produce a short numerical response, consistent with the system prompt instruction to respond with ``ONLY a single number from 1 to 10.''
 
Response parsing uses a regex-based extraction pipeline that identifies the first valid integer in the range 1--10 from the model's output text. In the rare event that parsing fails (e.g., the model produces text instead of a number), the query is retried up to 3 times before recording a null response. Across all 31{,}500 queries in the study, the refusal and parsing failure rate was \textbf{0\%}: every query produced a valid integer response. This perfect compliance rate likely reflects the simplicity of the constrained output format (a single digit) and the non-controversial framing of the question.
 
Checkpointing was implemented at 500-query intervals to enable fault-tolerant execution on the HPC cluster, where jobs are subject to wall-time limits and occasional preemption. The complete baseline evaluation of all three models required approximately 48.6 minutes of A100 GPU time.

\subsection{Statistical Analysis}
\label{sec:stats}
 
We employ three statistical tests corresponding to the three research questions, with a Bonferroni correction applied across the four hypothesis tests (2 models $\times$ 2 research questions for RQ2 and RQ3) to control the family-wise error rate at $\alpha = 0.05$.
 
\paragraph{RQ1: Cultural Origin Bias.} We test whether bias varies systematically across models and countries using a two-way analysis of variance (ANOVA) with model (3 levels: Bielik, Gemma3, Qwen) and country (3 levels: China, Slovakia, USA) as between-group factors, and $\hat{W}_1$ as the dependent variable. The $3 \times 3$ design yields 189 observations (63 personas $\times$ 3 models). We report main effects for model and country, as well as the model$\times$country interaction, which is of primary theoretical interest: a significant interaction indicates that different models exhibit different patterns of bias across countries, consistent with the hypothesis that cultural provenance shapes model behavior. Post-hoc pairwise comparisons use Tukey's Honestly Significant Difference (HSD) test.
 
\paragraph{RQ2: Fine-Tuning Efficacy.} For each fine-tuned model, we compare $\hat{W}_1$ scores before and after LoRA using a paired $t$-test on the 5 target personas. The pairing structure is natural: each persona serves as its own control, with baseline and post-LoRA $\hat{W}_1$ forming the paired observations. We report the mean change $\Delta \hat{W}_1$ (negative = improvement), Cohen's $d$ as a standardized effect size measure, and 95\% confidence intervals for the mean difference. The Bonferroni-corrected significance threshold is $\alpha / 4 = 0.0125$.
 
\paragraph{RQ3: Side Effects.} To assess whether fine-tuning introduces collateral damage, we perform a paired $t$-test on $\hat{W}_1$ scores for the 58 non-target personas (i.e., all personas except the 5 worst-case targets), again comparing baseline to post-LoRA values. However, because aggregate statistics can mask opposing effects that cancel out (a concern motivated by An et al.'s~\cite{an2025gender} finding that bias corrections can produce overcorrection in other demographic groups), we additionally decompose the side-effect analysis by country. This country-level decomposition, while not part of the formal hypothesis testing framework, reveals redistribution patterns that are invisible in the aggregate statistics and constitute one of the study's most important findings.
 
All statistical computations were performed in R. Bootstrap confidence intervals used the nonparametric percentile method with $B = 1{,}000$ resamples. Effect sizes are reported as Cohen's $d$ for paired designs, computed as the mean paired difference divided by the standard deviation of the paired differences. We follow standard conventions for interpreting $d$: small ($|d| < 0.5$), medium ($0.5 \leq |d| < 0.8$), and large ($|d| \geq 0.8$).

\section{Results}
 
This section presents findings for each research question in sequence. We first describe the baseline performance of all three models across 63 personas (Section~\ref{sec:baseline}), then test whether models favor their country of origin (Section~\ref{sec:rq1}), evaluate whether LoRA fine-tuning reduces bias on worst-case personas (Section~\ref{sec:rq2}), and finally assess side effects on non-target personas (Section~\ref{sec:rq3}).

\subsection{Baseline Model Performance}
\label{sec:baseline}
 
The baseline evaluation queried all three models across 63 personas with 100 responses each, generating 18,900 total responses. All queries returned valid integer responses in the range 1--10, yielding a 0\% refusal rate across all models and personas. The complete baseline evaluation required approximately 48.6 minutes of A100 GPU time.
 
\begin{figure*}[tp]
    \centering
    \includegraphics[width=\textwidth]{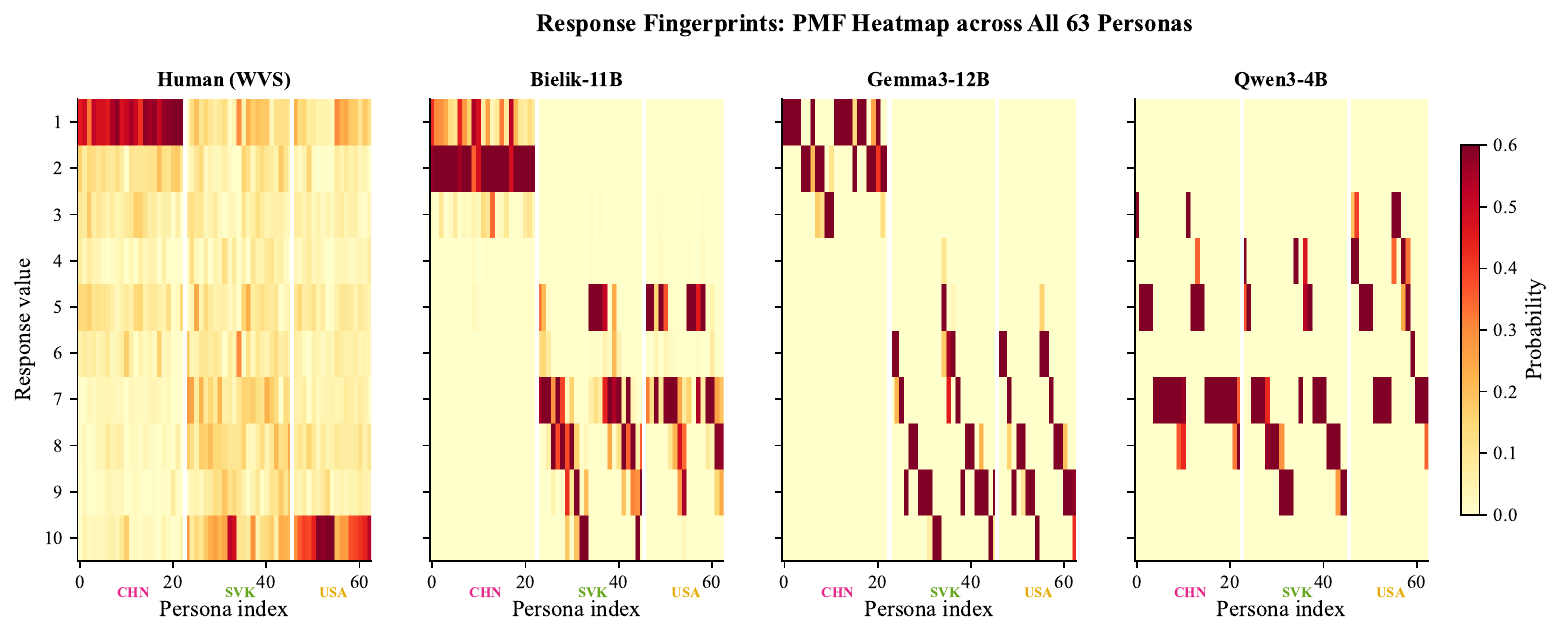}
    \caption{Response fingerprints: PMF heatmaps across all 63 personas for human WVS data and three models. Columns represent personas sorted by country (China, Slovakia, USA); rows represent response values (1--10); color intensity indicates probability mass. The human panel reveals three distinct national patterns. Models differ markedly in their ability to reproduce this structure.}
    \label{fig:fingerprint}
\end{figure*}

Figure~\ref{fig:fingerprint} provides a panoramic view of the response landscape. Each panel displays a probability mass function (PMF) heatmap for all 63 personas (columns, sorted by country: China, Slovakia, USA) across the 10-point response scale (rows), with darker cells indicating higher probability mass at that response value.
 
The human panel reveals three distinct national response patterns that reflect the cultural profiles summarized in Table~\ref{tab:country_profiles}. Chinese personas concentrate probability mass at values 1--2 (the secular pattern), Slovak personas exhibit a broad distribution with moderate mass across values 5--10, and American personas show a bimodal structure with a strong mode at 10 and a secondary concentration at 1. This heterogeneity establishes the ground truth that models must reproduce to achieve low $\hat{W}_1$.
 
The three models differ substantially in their ability to capture this structure:
 
\textbf{Bielik-11B} demonstrates the strongest persona differentiation among the three models. For Chinese personas, it correctly concentrates responses at low values (1--2), producing the characteristic secular fingerprint. For Slovak and American personas, it shifts toward moderate-to-high values (5--8), though it fails to reproduce the strong mode at 10 that characterizes American religiosity. This graded response pattern (low for China, moderate for Slovakia and the US) yields the lowest overall bias among the three models.
 
\textbf{Gemma3-12B} exhibits notably low response variance. Across nearly all 63 personas, regardless of country, the model concentrates probability mass at the value 7, producing a near-uniform fingerprint with minimal persona differentiation. This pattern is consistent with the behavior of quantized models served through inference frameworks like Ollama, where deterministic tendencies can dominate despite non-zero temperature settings. While the model's fixed response of 7 happens to be close to the means for Slovakia (6.4) and the United States (6.7), it is far from China's secular distribution (mean 2.8), yielding moderate $\hat{W}_1$ values for Chinese personas.
 
\textbf{Qwen3-4B} presents the most problematic pattern. Like Gemma3, it concentrates responses at high values (7--8) across most personas, but unlike Gemma3, it does so even for Chinese personas, where the correct response should be strongly concentrated at 1--2. This creates a striking paradox: a Chinese-origin model produces responses that are maximally misaligned with Chinese reality. We return to the interpretation of this finding in Section~\ref{sec:rq1}.

\subsection{Cultural Origin Bias Analysis (RQ1)}
\label{sec:rq1}
 
\noindent\textbf{RQ1: Do language models systematically favor demographic personas aligned with their country of origin?}
 
\medskip
 
To test whether cultural provenance shapes model bias, we conducted a two-way ANOVA with model (3 levels) and country (3 levels) as factors and $\hat{W}_1$ as the dependent variable ($N = 189$ observations). Table~\ref{tab:anova} presents the results.
 
\begin{table*}[tp]
    \centering
    \small
    \caption{Two-way ANOVA results for $\hat{W}_1$ as a function of model and persona country. All effects are statistically significant. The model$\times$country interaction confirms that bias patterns differ across models.}
    \label{tab:anova}
    \begin{tabular}{lrrrl}
        \toprule
        Source & df & $F$ & $p$ & Significance \\
        \midrule
        Model & 2, 180 & 56.54 & $<2 \times 10^{-16}$ & *** \\
        Country & 2, 180 & 5.25 & 0.006 & ** \\
        Model $\times$ Country & 4, 180 & 26.38 & $<2 \times 10^{-16}$ & *** \\
        \bottomrule
    \end{tabular}
\end{table*}
 
All three effects are statistically significant. The main effect of model ($F(2, 180) = 56.54$, $p < 2 \times 10^{-16}$) confirms that the three models differ substantially in their overall bias levels. The main effect of country ($F(2, 180) = 5.25$, $p = 0.006$) indicates that some countries are harder to align with than others. Most importantly, the significant model$\times$country interaction ($F(4, 180) = 26.38$, $p < 2 \times 10^{-16}$) demonstrates that models exhibit \emph{different patterns} of bias across countries, precisely the signature one would expect if cultural provenance matters.
 
\begin{figure*}[tp]
    \centering
    \includegraphics[width=0.85\textwidth]{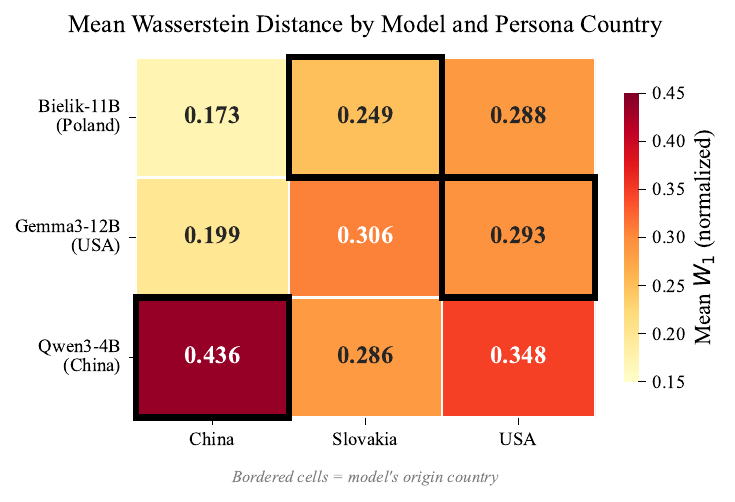}
    \caption{Mean Wasserstein distance by model origin and persona country. Bordered cells indicate the model's cultural origin country or closest available proxy. If models favored their origin, bordered cells would show the lowest values in each row. Instead, no model performs best on its origin-aligned population: Gemma and Bielik obtain their lowest distances on Chinese personas, while Qwen performs worst on Chinese personas and best on Slovak personas.}
    \label{fig:heatmap}
\end{figure*}

Table~\ref{tab:rq1_matrix} and Figure~\ref{fig:heatmap} present the mean $\hat{W}_1$ values for each model--country combination.
 
\begin{table*}[tp]
    \centering
    \small
    \caption{Mean normalized Wasserstein distance ($\hat{W}_1$) by model and persona country. Lower values indicate better alignment with human survey data. Bordered cells in Figure~\ref{fig:heatmap} indicate each model's cultural origin country.}
    \label{tab:rq1_matrix}
    \begin{tabular}{lccc}
        \toprule
        & China & Slovakia & USA \\
        \midrule
        Bielik-11B (Poland) & 0.173 & 0.249 & 0.288 \\
        Gemma3-12B (USA) & 0.199 & 0.306 & 0.293 \\
        Qwen3-4B (China) & 0.436 & 0.286 & 0.348 \\
        \bottomrule
    \end{tabular}
\end{table*}
 
Three key patterns emerge from this analysis:

\paragraph{Distribution-shape effect.} The results provide little support for a simple origin-country favoritism hypothesis. None of the three models obtains its lowest $\hat{W}_1$ on the population most directly associated with its country of origin. Gemma3-12B, despite being a US-origin model, performs slightly better on Chinese personas than on American ones. Bielik-11B, although trained in a Polish/Central European context, also obtains its lowest distance on Chinese personas rather than on the Slovak proxy population. This pattern is best understood as an artifact of distributional geometry rather than genuine cultural competence. China's response distribution on Q164 is heavily concentrated at the low end of the scale (52\% of respondents answer ``1''), making it a narrow, peaked target that a concentrated model response can partially approximate. In contrast, Slovakia and the United States have broader, more dispersed distributions that are harder to reproduce with the response patterns typically generated by language models. The lower distances for China in Bielik and Gemma therefore reflect the asymmetric difficulty of matching different distribution shapes, not a genuine preference for Chinese or secular values.

\paragraph{Limited regional-proximity pattern in Bielik.} Bielik-11B shows a weaker but more plausible regional-proximity pattern once the distribution-shape effect is taken into account. Among the two non-Chinese populations, Bielik aligns better with Slovakia ($\hat{W}_1 = 0.249$) than with the United States ($\hat{W}_1 = 0.288$). However, the evidence should be interpreted cautiously. The model still obtains its lowest distance on Chinese personas, and Slovakia is only a proxy for Poland rather than a direct Polish reference distribution. This suggests that distributional geometry can dominate cultural proximity, while regional proximity may only appear in comparisons between similarly dispersed non-Chinese populations.

\paragraph{Origin-country misalignment in Qwen.} The clearest counterexample to origin-country favoritism is Qwen3-4B, which achieves its \emph{worst} performance on Chinese personas ($\hat{W}_1 = 0.436$), the population most closely aligned with its training provenance. This is the highest $\hat{W}_1$ value in the entire 3$\times$3 matrix and is significantly different from all other model--country combinations in post-hoc Tukey HSD tests. The result is counterintuitive: a Chinese-origin model is the least capable of reproducing Chinese survey responses. As the fingerprint analysis in Figure~\ref{fig:fingerprint} reveals, Qwen consistently produces moderate-to-high religiosity responses even when the ground truth is strongly secular. This pattern is more consistent with a model-level response default, possibly shaped by instruction tuning or safety alignment, than with preservation of culturally specific Chinese opinion patterns. Proposed interpretation is consistent with the broader finding of Tao et al.~\cite{tao2024cultural} that models can exhibit culturally misaligned defaults regardless of the cultural context specified in the prompt.

\medskip

\noindent\textbf{Answer to RQ1:} Models do \emph{not} systematically favor their country of origin. Instead, baseline cultural alignment appears to be shaped by a combination of distributional difficulty, model-specific response defaults, and possible effects of instruction tuning. Bielik provides limited evidence of regional proximity only when comparing Slovakia with the United States, while Qwen provides a strong counterexample: a Chinese-origin model is least aligned with Chinese respondents.

\subsection{Fine-Tuning Efficacy (RQ2)}
\label{sec:rq2}
 
\noindent\textbf{RQ2: Does targeted LoRA fine-tuning on worst-case personas significantly reduce bias?}
 
\medskip
 
We applied LoRA fine-tuning to Bielik-11B and Qwen3-4B, targeting each model's five worst-case personas as identified by the bootstrap-based selection procedure (Section~\ref{sec:finetuning}). Bielik's targets were all American personas; Qwen's were all Chinese personas. After fine-tuning, each model was re-evaluated on all 63 personas. Table~\ref{tab:rq2} reports the before-and-after $\hat{W}_1$ values for each target persona.
 
\begin{table*}[tp]
    \centering
    \small
    \caption{RQ2: Fine-tuning efficacy on worst-case target personas. Negative $\Delta \hat{W}_1$ indicates improvement (bias reduction). Bielik achieved statistically significant improvement across all five targets; Qwen worsened on four of five.}
    \label{tab:rq2}
    \begin{tabular}{lccrc}
        \toprule
        Persona & $\hat{W}_1$ Before & $\hat{W}_1$ After & $\Delta$ & \% Change \\
        \midrule
        \multicolumn{5}{l}{\textit{Bielik-11B (targets: USA personas)}} \\
        \quad USA\_Male\_30-49\_Medium & 0.365 & 0.289 & $-$0.076 & $-$20.8\% \\
        \quad USA\_Male\_18-29\_Medium & 0.355 & 0.287 & $-$0.068 & $-$19.1\% \\
        \quad USA\_Male\_50-64\_Higher & 0.340 & 0.290 & $-$0.050 & $-$14.7\% \\
        \quad USA\_Female\_18-29\_Medium & 0.327 & 0.277 & $-$0.050 & $-$15.3\% \\
        \quad USA\_Male\_18-29\_Higher & 0.321 & 0.276 & $-$0.046 & $-$14.2\% \\
        \midrule
        \multicolumn{5}{l}{\textit{Qwen3-4B (targets: CHN personas)}} \\
        \quad CHN\_Male\_30-49\_Medium & 0.526 & 0.525 & $-$0.001 & $-$0.2\% \\
        \quad CHN\_Male\_50-64\_Lower & 0.531 & 0.622 & $+$0.091 & $+$17.1\% \\
        \quad CHN\_Male\_65+\_Higher & 0.611 & 0.703 & $+$0.092 & $+$15.0\% \\
        \quad CHN\_Male\_65+\_Lower & 0.583 & 0.723 & $+$0.140 & $+$24.0\% \\
        \quad CHN\_Male\_65+\_Medium & 0.637 & 0.723 & $+$0.086 & $+$13.5\% \\
        \bottomrule
    \end{tabular}
\end{table*}
 
\paragraph{Bielik-11B: successful bias reduction.} All five target personas showed reduced $\hat{W}_1$ after LoRA fine-tuning, with improvements ranging from $-14.2\%$ to $-20.8\%$ (mean: $-16.8\%$). The paired $t$-test confirms statistical significance even after Bonferroni correction: $t(4) = -9.811$, $p_{\text{raw}} = 0.000605$, $p_{\text{Bonferroni}} = 0.0024$, Cohen's $d = -4.388$ (very large effect), with a 95\% confidence interval for the mean difference of $[-0.074, -0.042]$.
 
This result is noteworthy because the LoRA adapter was trained exclusively on non-religious WVS questions (Section~\ref{sec:finetuning}), yet it improved alignment on the held-out religiosity question Q164. The adapter therefore learned transferable cultural patterns (how American personas of specific demographic profiles respond to survey questions in general) rather than memorizing the correct distribution for a single item. This transfer learning success validates the study's core methodological assumption that cross-question cultural knowledge can be leveraged to correct bias on held-out evaluation items.
 
\paragraph{Qwen3-4B: failed bias reduction.} In stark contrast, Qwen's LoRA fine-tuning \emph{increased} bias on four of five target personas, with worsening ranging from $+13.5\%$ to $+24.0\%$ (mean: $+13.9\%$). The only exception, CHN\_Male\_30-49\_Medium, showed negligible change ($-0.2\%$). While the aggregate worsening is substantial, the paired $t$-test does not reach significance after Bonferroni correction: $t(4) = 3.563$, $p_{\text{raw}} = 0.0235$, $p_{\text{Bonferroni}} = 0.094$, Cohen's $d = +1.593$. The raw $p$-value would be significant at $\alpha = 0.05$ without correction; the non-significance after Bonferroni adjustment reflects the conservative nature of the correction applied to a small sample of 5 paired observations, not the absence of a meaningful effect.
 
\begin{figure*}[tp]
    \centering
    \includegraphics[width=\textwidth]{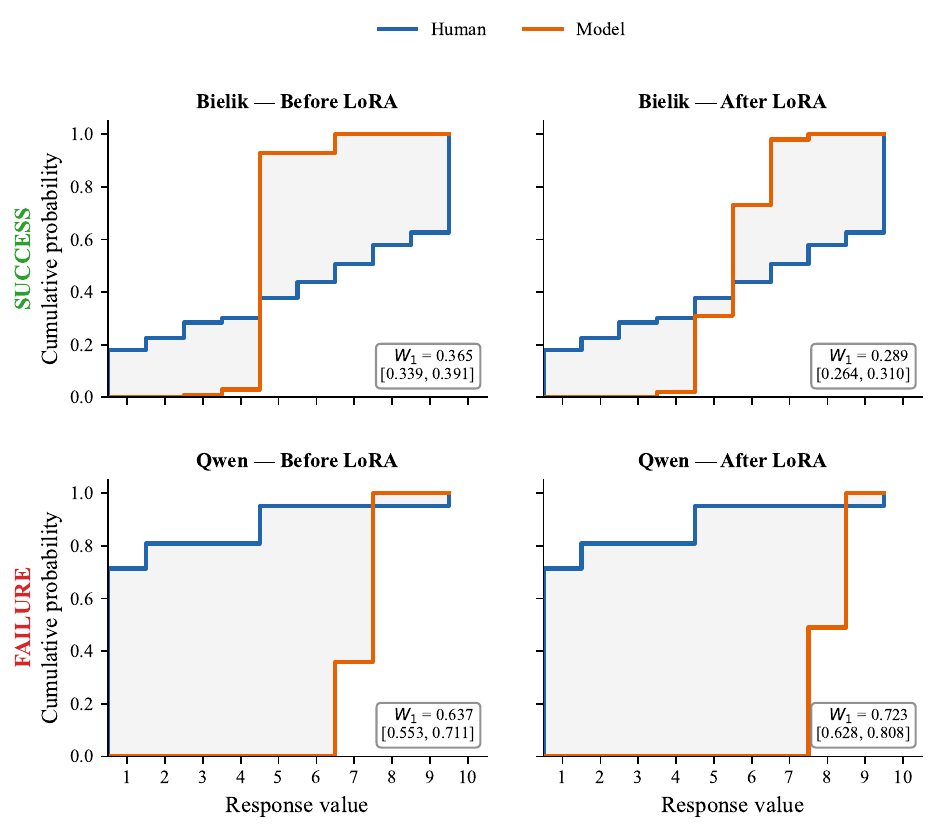}
    \caption{CDF overlays before and after LoRA fine-tuning for the worst-case target persona of each model. Top row: Bielik-11B on \texttt{USA\_Male\_30-49\_Medium}; the model CDF shifts toward the human CDF, reducing the shaded $W_1$ area from 0.365 to 0.289. Bottom row: Qwen3-4B on \texttt{CHN\_Male\_65+\_Medium}; the model CDF shifts \emph{further} from the human CDF, increasing $W_1$ from 0.637 to 0.723. Annotations show $\hat{W}_1$ with 95\% bootstrap confidence intervals.}
    \label{fig:cdf_comparison}
\end{figure*}

Figure~\ref{fig:cdf_comparison} visualizes the contrasting outcomes using CDF overlays for each model's most illustrative target persona.
 
In the Bielik panels (top row), the model CDF (orange) moves visibly closer to the human CDF (blue) after fine-tuning: the shaded area between the curves, representing $W_1$, shrinks from 0.365 [0.339, 0.391] to 0.289 [0.264, 0.310]. The model has learned to distribute more probability mass toward the high end of the scale, better reflecting American religiosity. In the Qwen panels (bottom row), the opposite occurs: the model CDF shifts \emph{further right} (toward even higher values), widening the gap with the steeply rising human CDF that reflects Chinese secularity. The $W_1$ increases from 0.637 [0.553, 0.711] to 0.723 [0.628, 0.808], indicating that the LoRA adapter amplified rather than corrected the existing bias.
 
\medskip
 
\noindent\textbf{Answer to RQ2:} LoRA fine-tuning on worst-case personas successfully reduces bias for Bielik-11B ($-16.8\%$, $p_{\text{Bonf}} = 0.0024$, $d = -4.4$) but backfires for Qwen3-4B ($+13.9\%$, $p_{\text{Bonf}} = 0.094$~n.s.). The asymmetry is consistent with model capacity differences (11B vs.\ 4B) and the interaction between LoRA corrections and existing RLHF alignment.

\subsection{Side Effects Analysis (RQ3)}
\label{sec:rq3}
 
\noindent\textbf{RQ3: Does fine-tuning to improve worst-case personas introduce negative side effects on other demographic groups?}
 
\medskip
 
For each fine-tuned model, we evaluate $\hat{W}_1$ changes on the 58 non-target personas (all personas except the 5 fine-tuning targets). Table~\ref{tab:rq3_aggregate} presents the aggregate results.
 
\begin{table*}[tp]
    \centering
    \small
    \caption{RQ3: Aggregate side effects on non-target personas. Neither model shows statistically significant aggregate change after Bonferroni correction.}
    \label{tab:rq3_aggregate}
    \begin{tabular}{lcrrrrc}
        \toprule
        Model & $n$ & Mean $\Delta \hat{W}_1$ & $t$ & df & $p_{\text{Bonf}}$ & Cohen's $d$ \\
        \midrule
        Bielik-11B & 58 & $+$0.019 & 1.775 & 57 & 0.325 & 0.233 \\
        Qwen3-4B & 58 & $+$0.012 & 1.244 & 57 & 0.875 & 0.163 \\
        \bottomrule
    \end{tabular}
\end{table*}
 
At the aggregate level, the answer to RQ3 appears straightforward: neither model shows statistically significant side effects after Bonferroni correction (Bielik: $p_{\text{Bonf}} = 0.325$; Qwen: $p_{\text{Bonf}} = 0.875$), and effect sizes are small (Cohen's $d = 0.233$ and $0.163$, respectively). If the analysis stopped here, one might conclude that LoRA fine-tuning is ``safe'': it can improve worst-case personas without degrading others.
 
However, aggregate statistics can conceal opposing effects that cancel in summation. Motivated by the overcorrection patterns documented by An et al.~\cite{an2025gender}, we decompose the side-effect analysis by persona country. Table~\ref{tab:rq3_country} presents this decomposition, which reveals a \emph{bias redistribution} pattern that the aggregate analysis completely obscures.
 
\begin{table*}[tp]
    \centering
    \small
    \caption{RQ3: Country-level decomposition of side effects on non-target personas. The aggregate non-significance masks a dramatic redistribution: Bielik's LoRA training on USA personas devastates Chinese personas ($+50.2\%$) while improving Slovak personas ($-10.4\%$). Qwen's training on CHN personas improves American personas ($-13.5\%$) while further worsening Chinese non-targets ($+27.6\%$).}
    \label{tab:rq3_country}
    \begin{tabular}{llcrrrr}
        \toprule
        Model & Country & $n$ & Mean $\Delta \hat{W}_1$ & \% Change & Improved & Worsened \\
        \midrule
        \multirow{3}{*}{Bielik-11B} 
            & China & 23 & $+$0.077 & $+$50.2\% & 3 & 20 \\
            & Slovakia & 23 & $-$0.032 & $-$10.4\% & 20 & 3 \\
            & USA (non-target) & 12 & $+$0.008 & $+$6.3\% & 6 & 6 \\
        \midrule
        \multirow{3}{*}{Qwen3-4B}
            & China (non-target) & 18 & $+$0.085 & $+$27.6\% & 2 & 16 \\
            & Slovakia & 23 & $-$0.001 & $+$1.6\% & 9 & 10 \\
            & USA & 17 & $-$0.047 & $-$13.5\% & 15 & 1 \\
        \bottomrule
    \end{tabular}
\end{table*}
 
\paragraph{Bielik: the redistribution effect.} Bielik's LoRA adapter, trained on five American personas with moderate-to-high religiosity (WVS mean $\approx 6.7$), shifts the model's response distribution \emph{globally} toward higher values. This produces three distinct outcomes depending on the cultural profile of the evaluated persona:
 
Chinese personas ($+50.2\%$ mean worsening, 20 of 23 worsened): The global upward shift pushes responses away from the low values that characterize Chinese secularity. This is the most severe collateral damage: the model sacrifices alignment with one culture to gain alignment with another.
 
Slovak personas ($-10.4\%$ mean improvement, 20 of 23 improved): Because Slovakia's religiosity profile (mean 6.4) is similar to the American targets (mean 6.7), the same upward shift that improves American alignment also benefits Slovak personas. This ``free improvement'' is an incidental consequence of cultural proximity.
 
Non-target American personas ($+6.3\%$ mean, 6 improved and 6 worsened): The remaining American personas show mixed results, suggesting that the LoRA adapter's corrections are not perfectly generalizable even within the target country.
 
\begin{figure*}[tp]
    \centering
    \includegraphics[width=\textwidth]{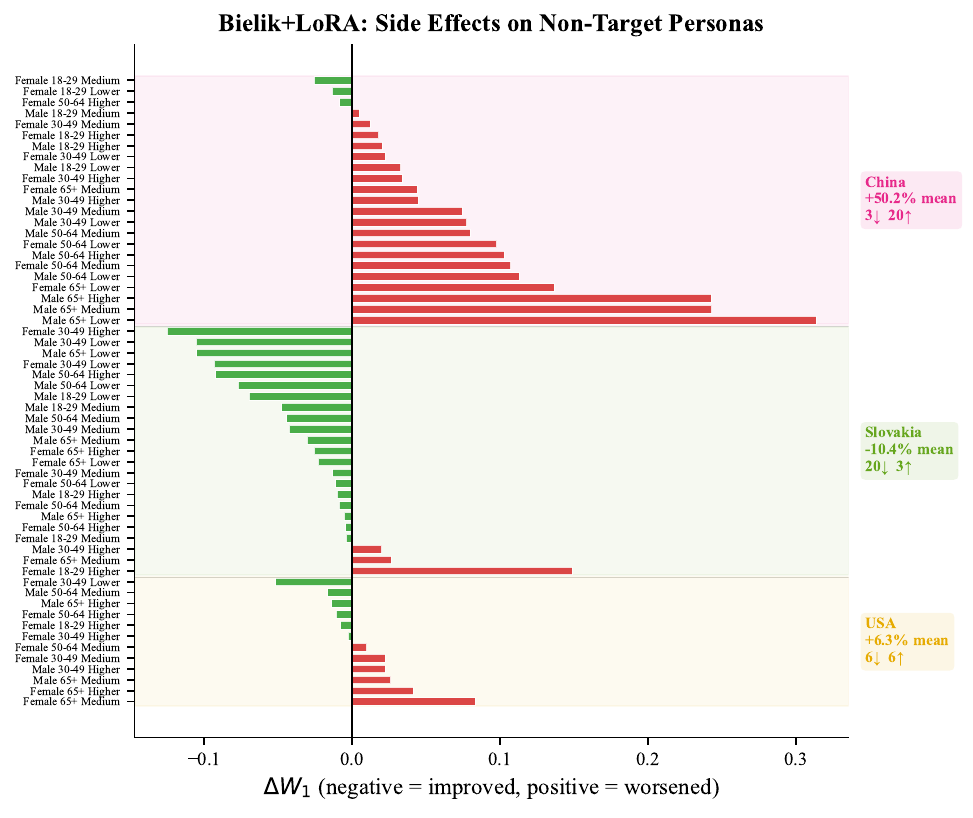}
    \caption{Bielik+LoRA: side effects on all 58 non-target personas, grouped by country. Bars extending right (red) indicate increased $\hat{W}_1$ (worsened bias); bars extending left (green) indicate decreased $\hat{W}_1$ (improved bias). Training on American personas devastated Chinese personas ($+50.2\%$ mean) while incidentally improving Slovak personas ($-10.4\%$ mean). The Qwen redistribution pattern is presented in Appendix~A.}
    \label{fig:redistribution_bielik}
\end{figure*}

Figure~\ref{fig:redistribution_bielik} provides a persona-level visualization of this redistribution. The horizontal diverging bars, grouped by country, make the pattern unmistakable: nearly every Chinese persona (pink region) extends to the right (worsened), while nearly every Slovak persona (green region) extends to the left (improved).
 
The most extreme collateral damage occurred among elderly Chinese personas. CHN\_Male\_65+\_Lower experienced a $+251\%$ increase in $\hat{W}_1$, making it Bielik's new single worst-case persona after fine-tuning. More broadly, Bielik's worst-case ranking underwent a \emph{complete turnover}: before fine-tuning, all five worst-case personas were American; after fine-tuning, all five are Chinese elderly personas (CHN\_Male\_65+\_Lower, $\hat{W}_1 = 0.439$; CHN\_Male\_65+\_Medium, 0.400; CHN\_Male\_65+\_Higher, 0.374; CHN\_Female\_65+\_Lower, 0.356; CHN\_Female\_50-64\_Lower, 0.308). The overlap between pre- and post-LoRA worst-case sets is zero out of five, a complete swap from one cultural group to another.
 
\paragraph{Qwen: a mirror-image redistribution.} Qwen's LoRA adapter, trained on five Chinese personas, produces a redistribution pattern that mirrors Bielik's in an instructive way. American personas improve by $-13.5\%$ (15 of 17 improved), while Chinese non-target personas worsen by $+27.6\%$ (16 of 18 worsened). Slovak personas show negligible change ($+1.6\%$). However, because Qwen's fine-tuning failed to improve even its target personas (Section~\ref{sec:rq2}), the redistribution here compounds failure rather than creating a meaningful tradeoff: Chinese personas worsen across the board, both targets and non-targets.
 
Qwen's worst-case ranking shows 4 of 5 overlap between pre- and post-LoRA sets: the same Chinese elderly personas remain at the top, now with even higher $\hat{W}_1$ values. This stability confirms that the LoRA adapter was unable to dislodge the deeply entrenched bias patterns in the smaller model.
 
\begin{figure*}[tp]
    \centering
    \includegraphics[width=\textwidth]{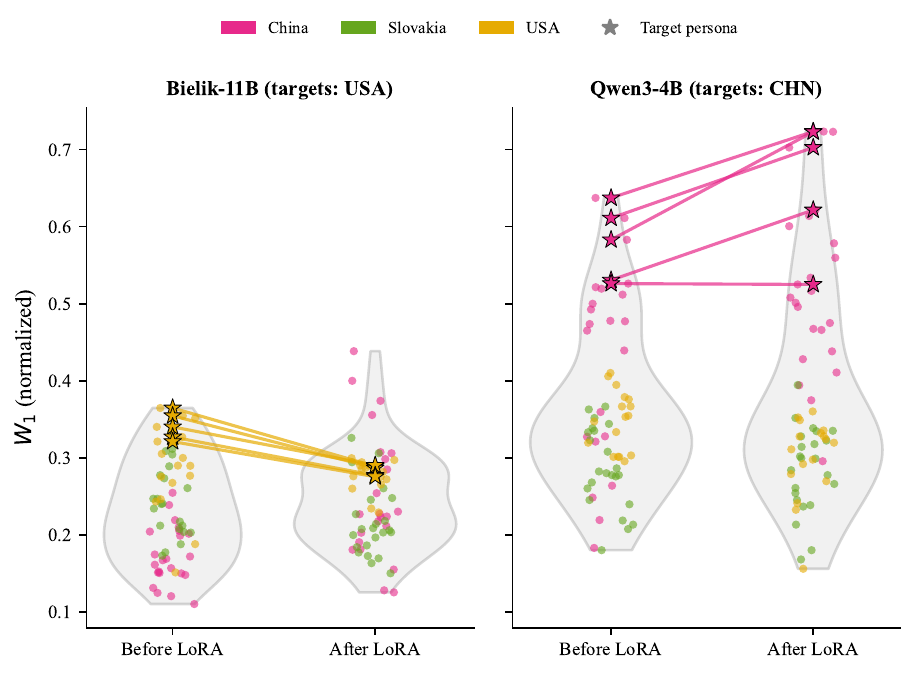}
    \caption{Before-and-after $\hat{W}_1$ distributions for all 63 personas, by model. Each dot represents one persona, colored by country (pink = China, green = Slovakia, yellow = USA). Star markers with connecting lines indicate the five target personas. Left panel (Bielik): targets drop from 0.32--0.37 to 0.28--0.29; Chinese non-targets shift upward; Slovak non-targets shift downward. Right panel (Qwen): targets rise from 0.53--0.64 to 0.52--0.72; American non-targets shift downward.}
    \label{fig:violin}
\end{figure*}

\paragraph{Combined visualization.} Figure~\ref{fig:violin} presents a violin-and-dot plot that captures the full before-and-after picture for both models simultaneously. Each dot represents one persona, colored by country; star markers with connecting lines highlight the five target personas.
 
In the left panel (Bielik), the target personas (yellow stars, USA) descend in a tight cluster from the upper portion of the violin to a narrow band around 0.28--0.29, confirming the uniform effectiveness of the fine-tuning. Meanwhile, the overall distribution compresses slightly, but the country-coded dots reveal the redistribution: pink dots (China) drift upward while green dots (Slovakia) drift downward. In the right panel (Qwen), the target personas (pink stars, China) ascend, their lines sloping upward and confirming the backfire effect, while yellow dots (USA) shift downward, reflecting the mirror-image redistribution.
 
\paragraph{Summary of statistical tests.} Table~\ref{tab:summary_tests} consolidates the results of all four hypothesis tests with Bonferroni correction.
 
\begin{table*}[tp]
    \centering
    \small
    \caption{Summary of all statistical tests (RQ2 and RQ3). Bonferroni correction applied across 4 tests ($\alpha_{\text{adjusted}} = 0.0125$). Only Bielik's RQ2 result reaches significance.}
    \label{tab:summary_tests}
    \begin{tabular}{llrrrrc}
        \toprule
        RQ & Model & Mean $\Delta \hat{W}_1$ & $t$ & $p_{\text{Bonf}}$ & Cohen's $d$ & Sig. \\
        \midrule
        RQ2 (targets) & Bielik & $-$0.058 & $-$9.811 & 0.0024 & $-$4.388 & *** \\
        RQ2 (targets) & Qwen & $+$0.082 & 3.563 & 0.094 & $+$1.593 & n.s. \\
        RQ3 (non-targets) & Bielik & $+$0.019 & 1.775 & 0.325 & 0.233 & n.s. \\
        RQ3 (non-targets) & Qwen & $+$0.012 & 1.244 & 0.875 & 0.163 & n.s. \\
        \bottomrule
    \end{tabular}
\end{table*}
 
\medskip
 
\noindent\textbf{Answer to RQ3:} No statistically significant aggregate side effects are observed for either model. However, country-level decomposition reveals a critical \emph{bias redistribution} pattern: LoRA fine-tuning does not remove bias so much as move it between cultural groups. Training on one country's personas improves alignment for culturally similar populations while degrading alignment for culturally distant ones. Aggregate metrics alone are insufficient to detect this redistribution, underscoring the need for subgroup-level evaluation in any bias mitigation pipeline.

\section{Discussion and Conclusions}
\label{sec:discussion}

\subsection{Interpretation of Findings}

The results of this study reveal a richer and more paradoxical picture of cultural bias in LLMs than prior work has suggested. Rather than a simple narrative of models favoring their cultural origins, we observe three distinct phenomena: a distribution-shape effect, an origin-country misalignment, and a bias redistribution effect, each of which demands a separate explanation.

\paragraph{Distribution-shape effect.} The finding that Gemma3-12B and Bielik-11B achieve their lowest $\hat{W}_1$ on Chinese personas (Section~\ref{sec:rq1}) might initially suggest stronger cultural alignment with Chinese respondents. However, this interpretation is misleading. China's response distribution for Q164 is heavily concentrated at the low end of the scale, with 52\% of respondents answering ``1.'' A distribution this peaked is geometrically easier to approximate: even a crude model that places most of its probability mass in the range 1--3 can achieve a relatively low $\hat{W}_1$. In contrast, Slovakia and the United States have broader, more dispersed distributions, bimodal in the case of the US, that require the model to spread probability mass across a wider range of values. The apparent advantage on Chinese personas therefore reflects an asymmetry in the \emph{difficulty of matching different distribution shapes}, not genuine cultural competence. This insight has methodological implications: cross-cultural bias comparisons must account for the intrinsic difficulty of each target distribution, and aggregate metrics can be misleading when target distributions differ substantially in their concentration.

\paragraph{The Qwen paradox: RLHF as bias amplifier.} Qwen3-4B's dramatic misalignment with its own origin country ($\hat{W}_1 = 0.436$ on Chinese personas, the highest value in the 3$\times$3 matrix) is the study's most counterintuitive finding. As the fingerprint analysis (Figure~\ref{fig:fingerprint}) reveals, Qwen systematically produces moderate-to-high religiosity responses (7--8) even when the ground truth is strongly secular. We interpret this as an artifact of the RLHF alignment process. When human annotators evaluate candidate model responses to a question about the importance of God, moderate-to-positive answers are likely rated as ``safer'' and more ``appropriate'' than answers at the extreme low end, which might be perceived as dismissive or culturally insensitive. The reward model trained on these annotations would then steer the LLM away from low-religiosity outputs, precisely the outputs needed to align with Chinese reality. This mechanism is consistent with Rozado's~\cite{rozado2024politicalpreferencesllms} finding that RLHF crystallizes specific value preferences during post-training alignment, and with the broader observation of Tao et al.~\cite{tao2024cultural} that all models they tested gravitate toward Western-aligned cultural values regardless of the specified cultural context. In effect, RLHF functions as a bias \emph{amplifier} on culturally sensitive topics, converting an initially neutral or mildly skewed base model into one that actively suppresses culturally authentic but politically inconvenient responses.

\paragraph{Bias redistribution: global shift, not targeted correction.} The country-level decomposition of side effects (Table~\ref{tab:rq3_country}) reveals that LoRA fine-tuning does not remove bias in an absolute sense but rather \emph{redistributes} it across cultural groups. The mechanism is straightforward: the LoRA adapter learns a directional correction that pushes responses toward higher values (Bielik, trained on religious American personas) or lower values (Qwen, trained on secular Chinese personas) and applies it globally across all personas. The adapter cannot distinguish ``this specific persona needs higher values'' from ``all personas need higher values,'' because the low-rank parameter update affects the model's attention layers indiscriminately.

\begin{figure*}[tp]
    \centering
    \includegraphics[width=\textwidth]{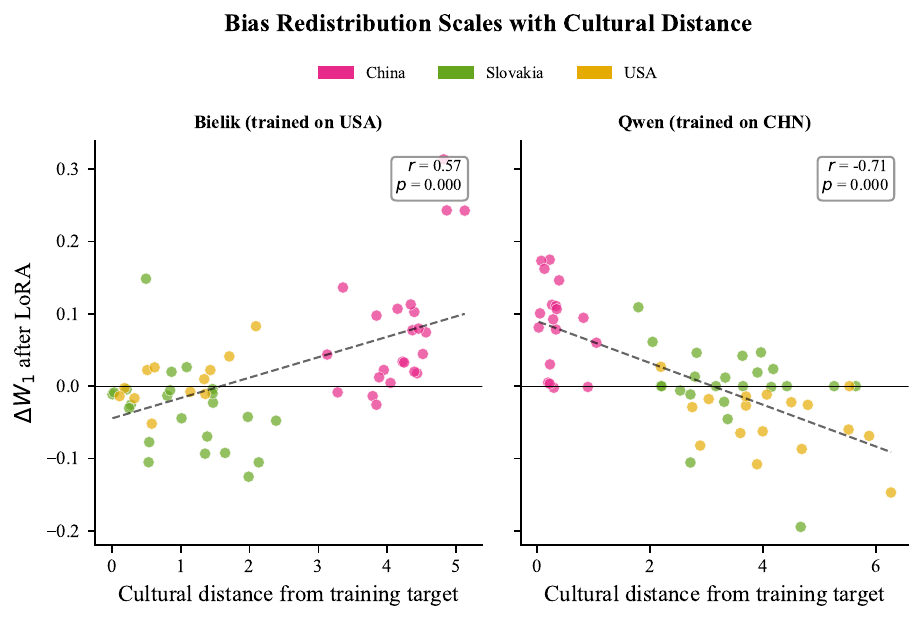}
    \caption{Bias redistribution scales with cultural distance from the fine-tuning target. Each dot represents a non-target persona, colored by country. The $x$-axis measures the absolute difference between the persona's country mean on Q164 and the training target's country mean. Left: Bielik ($r = 0.57$); personas distant from American religiosity (i.e., Chinese secular personas) worsen most. Right: Qwen ($r = -0.71$); personas distant from Chinese secularity (i.e., American personas) improve most, despite the fine-tuning failing on its actual targets.}
    \label{fig:cultural_distance}
\end{figure*}

Figure~\ref{fig:cultural_distance} quantifies this redistribution. The scatter plots show $\Delta \hat{W}_1$ as a function of cultural distance from the fine-tuning target, measured as the absolute difference between each persona's WVS country mean and the training target's mean. For Bielik ($r = 0.57$, $p < 0.001$), the correlation is positive: personas culturally distant from the American training targets (i.e., Chinese personas) experience the greatest worsening, while culturally proximate personas (Slovak) experience incidental improvement. For Qwen ($r = -0.71$, $p < 0.001$), the correlation is negative: the attempted downward shift toward secular responses, which failed for the Chinese targets due to RLHF resistance, succeeded in improving alignment for the already-religious American personas. These correlations demonstrate that the magnitude of side effects is not random but scales systematically with cultural distance from the fine-tuning targets.

\begin{figure*}[tp]
    \centering
    \includegraphics[width=0.85\textwidth]{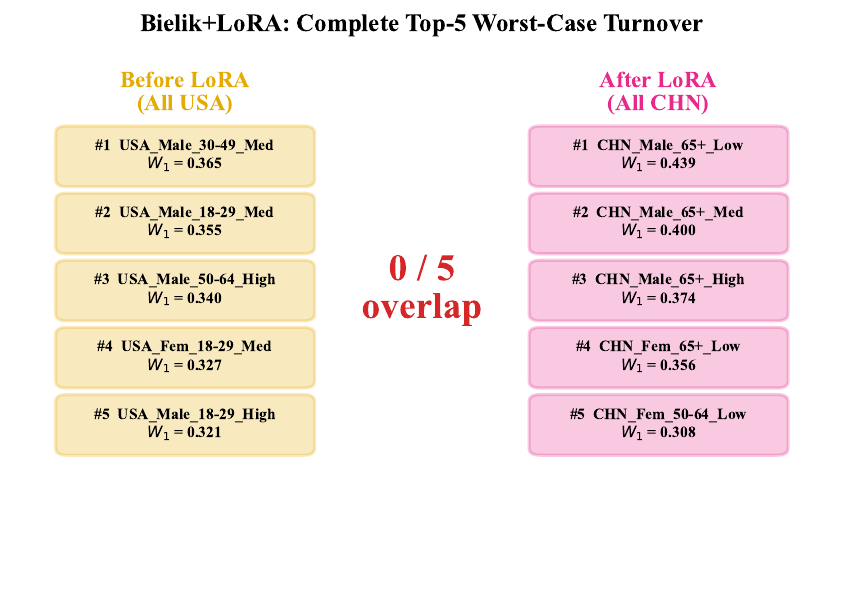}
    \caption{Bielik+LoRA: complete worst-case turnover. Before fine-tuning, all five worst-case personas were American; after fine-tuning, all five are Chinese elderly. The overlap between pre- and post-LoRA worst-case sets is zero, a complete cultural swap.}
    \label{fig:turnover}
\end{figure*}

This redistribution creates a ``whack-a-mole'' dynamic: fixing bias for one cultural group creates new worst-case personas in another. Figure~\ref{fig:turnover} illustrates this for Bielik, where the five worst-case personas swap entirely from American to Chinese elderly between the pre- and post-LoRA evaluations, with zero overlap. This finding connects directly to the overcorrection patterns documented by An et al.~\cite{an2025gender} in the context of gender and racial bias, and suggests that targeted fine-tuning without counterfactual constraints on non-target groups is fundamentally limited as a bias mitigation strategy.

\paragraph{Model scale and LoRA responsiveness.} The divergent fine-tuning outcomes (Bielik-11B succeeded while Qwen3-4B failed) are consistent with the emerging evidence that model scale affects receptiveness to parameter-efficient adaptation. Hu and Collier~\cite{hu2024personaeffect} found that smaller models (7B and 13B) performed substantially worse than 70B models at utilizing persona information, capturing almost none of the target variance at the 7B scale. Liu et al.~\cite{liu2025alignment} reported monotonically increasing alignment scores with model size across the Qwen2.5 family. Haller et al.~\cite{haller2024opiniongpt} observed that larger base models captured demographic-specific biases more precisely with LoRA adapters. Bai et al.~\cite{bai2025biased} found that larger models retained richer implicit associations despite explicit debiasing, suggesting greater representational capacity that fine-tuning can leverage. Our results are consistent with this pattern: the 11B model had sufficient representational capacity to integrate cultural signals from 1,180 training pairs, while the 4B model did not. However, we cannot fully disentangle model scale from other confounds: Qwen's stronger RLHF alignment, fewer training pairs (1,003 vs.\ 1,180), and the greater baseline severity of its misalignment ($\hat{W}_1$ 0.53--0.64 vs.\ 0.32--0.37 for Bielik) may all have contributed to the failure.

\subsection{Practical Implications}

Our findings suggest a three-step framework for model developers conducting pre-deployment cultural bias assessment:

\textbf{Step 1: Identify.} Compute $\hat{W}_1$ between model response distributions and human reference populations for a comprehensive set of demographic personas. Use bootstrap confidence intervals and select worst-case personas by the lower CI bound to ensure robust identification.

\textbf{Step 2: Mitigate.} Apply targeted LoRA fine-tuning to the worst-case personas using training data from non-overlapping survey items to prevent data leakage. The modest computational requirements (under 15 minutes on a single A100, fewer than 1,200 training pairs) make this feasible as a routine component of model evaluation.

\textbf{Step 3: Evaluate at the subgroup level.} Crucially, assess side effects not only in aggregate but decomposed by country, demographic group, and other relevant subpopulations. Our results demonstrate that aggregate non-significance can mask severe redistribution effects invisible to summary statistics.

This framework is designed to be reproducible and accessible: it requires only open-weight models deployable on a single GPU, publicly available survey data (WVS), and standard statistical tools. All code and data from this study are publicly available to facilitate adoption. We emphasize that this methodology is intended as a \emph{diagnostic} tool that identifies where models fail, rather than a complete solution to cultural bias. The redistribution problem (Section~\ref{sec:rq3}) demonstrates that targeted LoRA fine-tuning in its current form is insufficient for comprehensive bias correction and must be complemented by constraint-aware training approaches.

\subsection{Limitations}

Several limitations constrain the generalizability of our findings.

\textbf{Single evaluation question.} Our entire analysis rests on WVS Q164 (``Importance of God''), a single item probing religiosity. While Q164 exhibits maximum cross-cultural variance and serves well as a proof of concept, we cannot claim that the observed patterns, such as the Qwen paradox or the redistribution effect, generalize to other value dimensions such as political orientation, gender attitudes, or economic values. The interaction between RLHF alignment and specific question content may produce different bias patterns for different topics.

\textbf{Three countries.} China, Slovakia, and the United States represent only a fraction of the world's cultural diversity. Expanding to additional countries would test whether our findings hold across a broader cultural spectrum, particularly countries from sub-Saharan Africa, South Asia, and the Middle East, regions consistently underserved in prior alignment studies~\cite{liu2025alignment}.

\textbf{Slovakia as proxy for Poland.} While Slovakia and Poland share substantial cultural proximity as Visegrad Group members with parallel post-socialist trajectories and comparable religiosity profiles, they are not identical. Slovak WVS distributions may not perfectly represent the opinions of the Polish population against which Bielik's training data is calibrated. Future work should replicate this analysis using Polish data when it becomes available in subsequent WVS waves.

\textbf{Prompt sensitivity.} Our study uses a single prompt template for all queries. Ceron et al.~\cite{ceron2024political} demonstrated that model responses to opinion questions are sensitive to prompt rephrasing and answer choice ordering. We did not test robustness to prompt variations, and it is possible that different prompting strategies would yield different $\hat{W}_1$ values.

\textbf{Quantized model behavior.} Gemma3-12B's low response variance (Figure~\ref{fig:fingerprint}) may partly reflect the deterministic tendencies of models served through the Ollama inference framework rather than an inherent property of the model architecture. This limits the interpretability of Gemma3's baseline results, although it does not affect the LoRA analysis (which excludes Gemma3).

\textbf{Intersectional effects.} An et al.~\cite{an2025gender} showed that bias patterns at demographic intersections (e.g., race $\times$ gender) can differ qualitatively from patterns observed along individual dimensions. Our persona design crosses four features but does not systematically analyze higher-order interactions, which may reveal additional bias patterns invisible to the current analysis.

\subsection{Future Work}

Several directions emerge naturally from the limitations and findings of this study.

\textbf{Multi-question evaluation.} Extending the analysis from a single religiosity item to dozens of WVS questions spanning political values, gender attitudes, economic preferences, and social trust would test whether redistribution effects generalize across value dimensions or are specific to the religiosity domain.

\textbf{Geographic expansion.} Scaling from 3 to 10 or more countries, selected to span the full range of the Inglehart-Welzel World Cultural Map, would provide a more rigorous test of the cultural distance hypothesis and identify regions where model bias is most severe.

\textbf{Constraint-based LoRA.} The redistribution problem motivates a modified fine-tuning objective that penalizes degradation on non-target populations. One approach would add a regularization term to the LoRA training loss that measures $\hat{W}_1$ on a held-out set of non-target personas, creating an explicit multi-objective optimization that balances targeted improvement against collateral damage.

\textbf{DPO comparison.} Direct Preference Optimization~\cite{jinnai2024morality} offers an alternative to SFT-based LoRA that operates on preference pairs rather than single-best responses. Comparing DPO and LoRA on the same worst-case personas would clarify whether the redistribution problem is specific to SFT training or inherent to parameter-efficient methods more broadly.

\textbf{Longitudinal analysis.} The WVS spans seven waves of data collection from 1981 to 2022. Tracking how model-population alignment evolves across waves would reveal whether LLMs reflect a static cultural snapshot or adapt (through retraining) to shifting societal values.

\textbf{Intersectional analysis.} Systematically testing interactions between demographic features would extend the persona-level analysis into a fully intersectional framework, asking, for example, whether elderly Chinese women are more poorly served than young Chinese men beyond what the marginal effects of age and sex would predict.

\subsection{Concluding Remarks}

This study demonstrates that cultural bias in open-weight LLMs is not only measurable but exhibits systematic and interpretable patterns. We established that the model$\times$country interaction is highly significant ($p < 2 \times 10^{-16}$), confirming that cultural provenance shapes model behavior in statistically robust ways. We showed that targeted LoRA fine-tuning on just five worst-case personas reduced bias by 16.8\% in Bielik-11B ($d = -4.4$), with all five targets improving, using fewer than 1,200 training pairs and under 15 minutes of computation on a single GPU. And we uncovered a striking paradox in Qwen3-4B: a Chinese-origin model that performs worst on its own Chinese population ($\hat{W}_1 = 0.436$, the highest value in the entire model$\times$country matrix), providing direct empirical evidence that safety alignment can actively suppress culturally authentic responses rather than preserve them.

Perhaps most valuable is the country-level decomposition methodology, which revealed the complete worst-case turnover effect and redistribution dynamics entirely invisible to aggregate statistics, offering the research community a new analytical lens for evaluating bias mitigation strategies. To our knowledge, this is the first study to combine worst-case persona identification with targeted LoRA fine-tuning for cross-cultural bias mitigation. All code, data, and trained adapters are publicly released, providing a reproducible benchmark that any research team can replicate on a single consumer GPU.

\begin{acknowledgment}
This work utilized the Alpine high performance computing resource at the University of Colorado Boulder. Alpine is jointly funded by the University of Colorado Boulder, the University of Colorado Anschutz, and Colorado State University, with support from NSF grants OAC-2201538 and OAC-2322260~\cite{alpine2023}. This research used data from the World Values Survey Wave~7 (2017--2022).\footnote{\url{https://www.worldvaluessurvey.org}} The complete codebase and data are available online.\footnote{\url{https://github.com/AntoniCzolgowski/llm-cultural-bias}}
\end{acknowledgment}

\bibliographystyle{plainurl}
\bibliography{references}

\clearpage
\onecolumn
\raggedbottom

\appendix
\section{Supplementary Figures}

\begin{figure}[H]
    \centering
    \includegraphics[width=\textwidth]{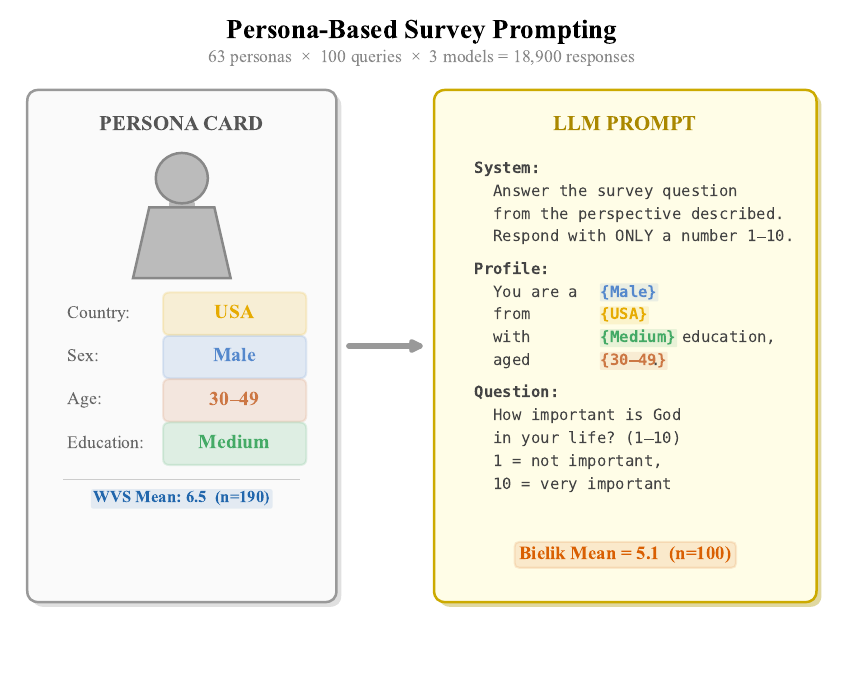}
    \caption{Persona-based survey prompting pipeline. A demographic persona card from the WVS is translated into a structured LLM prompt consisting of a system instruction, a demographic profile, and the survey question. Each persona is queried 100 times to construct an empirical response distribution.}
    \label{fig:prompt_template}
\end{figure}

\begin{figure}[H]
    \centering
    \includegraphics[width=\textwidth]{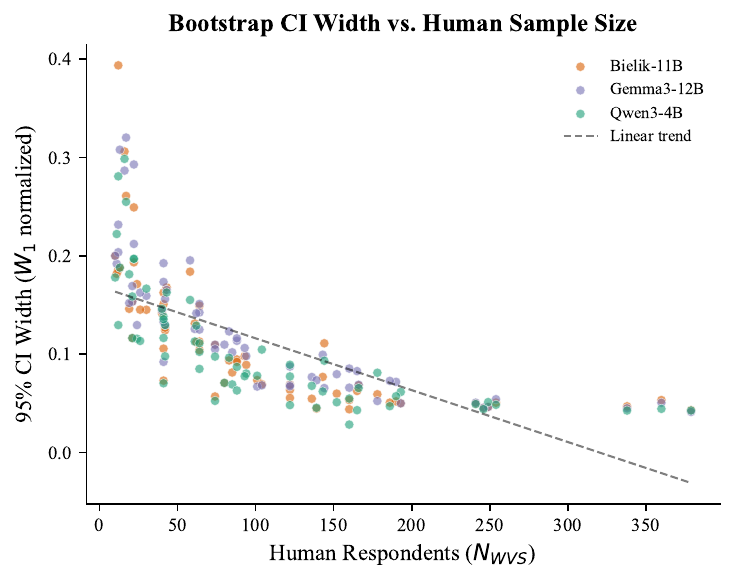}
    \caption{Bootstrap 95\% confidence interval width as a function of human sample size ($N_{\text{WVS}}$). CI width decreases monotonically with sample size, confirming that the bootstrap procedure produces tighter estimates for personas with more WVS respondents. The dashed line shows the linear trend.}
    \label{fig:ci_vs_n}
\end{figure}

\begin{figure}[H]
    \centering
    \includegraphics[width=\textwidth]{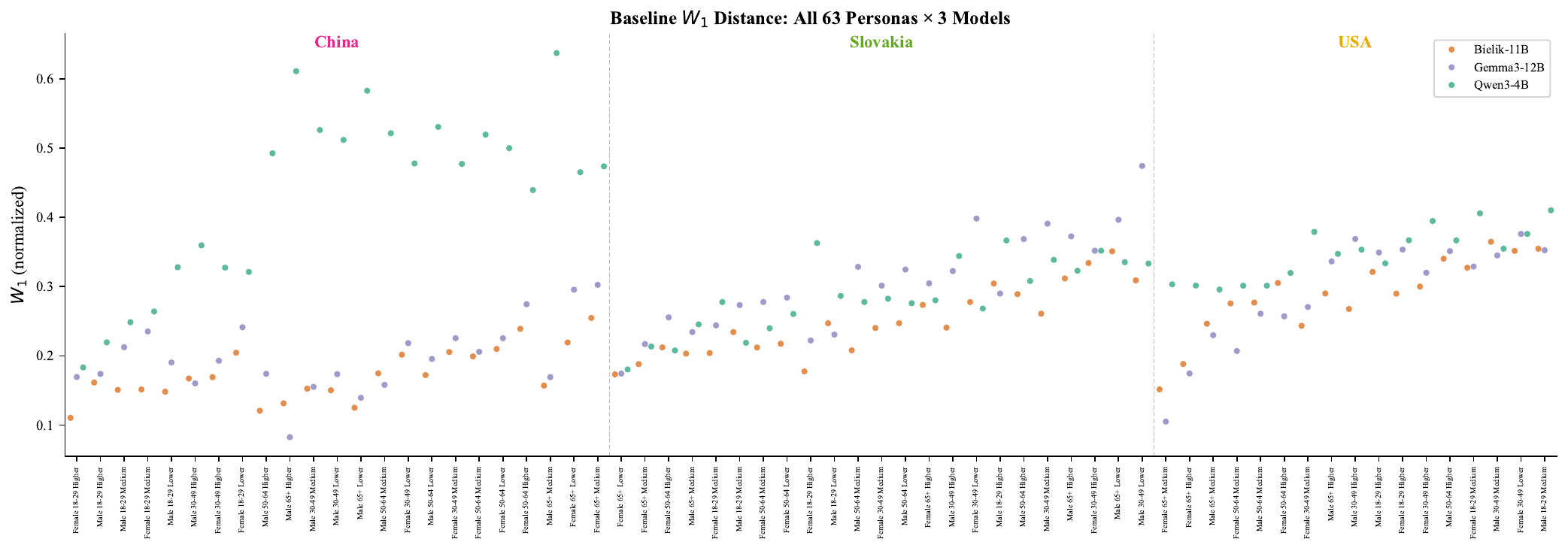}
    \caption{Baseline $\hat{W}_1$ distances for all 63 personas across three models. Each dot represents one persona, colored by country. Personas are sorted within each model panel by $\hat{W}_1$. Qwen3-4B exhibits the widest spread and the highest individual values, concentrated among Chinese personas.}
    \label{fig:strip_all}
\end{figure}

\begin{figure}[H]
    \centering
    \includegraphics[width=\textwidth]{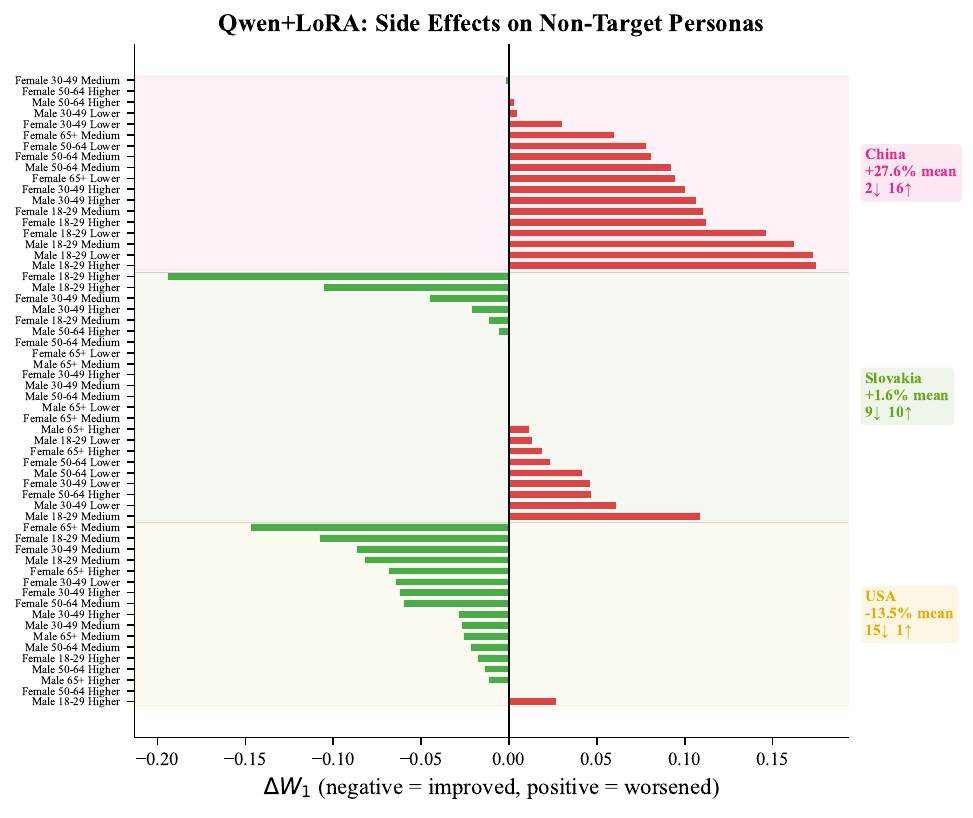}
    \caption{Qwen+LoRA: side effects on all 58 non-target personas, grouped by country. Compare with Figure~\ref{fig:redistribution_bielik} for Bielik. The mirror-image pattern is evident: Chinese non-targets worsen ($+27.6\%$ mean, 16 of 18 worsened) while American personas improve ($-13.5\%$ mean, 15 of 17 improved). Slovak personas show negligible aggregate change.}
    \label{fig:redistribution_qwen}
\end{figure}

\begin{figure}[H]
    \centering
    \includegraphics[width=0.75\textwidth]{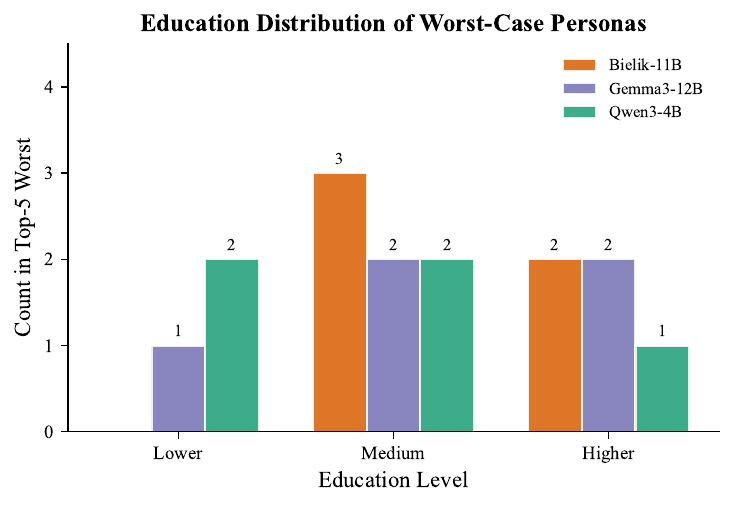}
    \caption{Education level distribution of worst-case personas across the three models. Medium and higher education levels dominate the worst-case lists, contrary to the intuitive expectation that lower-education personas, typically underrepresented in LLM training data, would be most poorly served.}
    \label{fig:education}
\end{figure}

\begin{figure}[H]
    \centering
    \includegraphics[width=\textwidth]{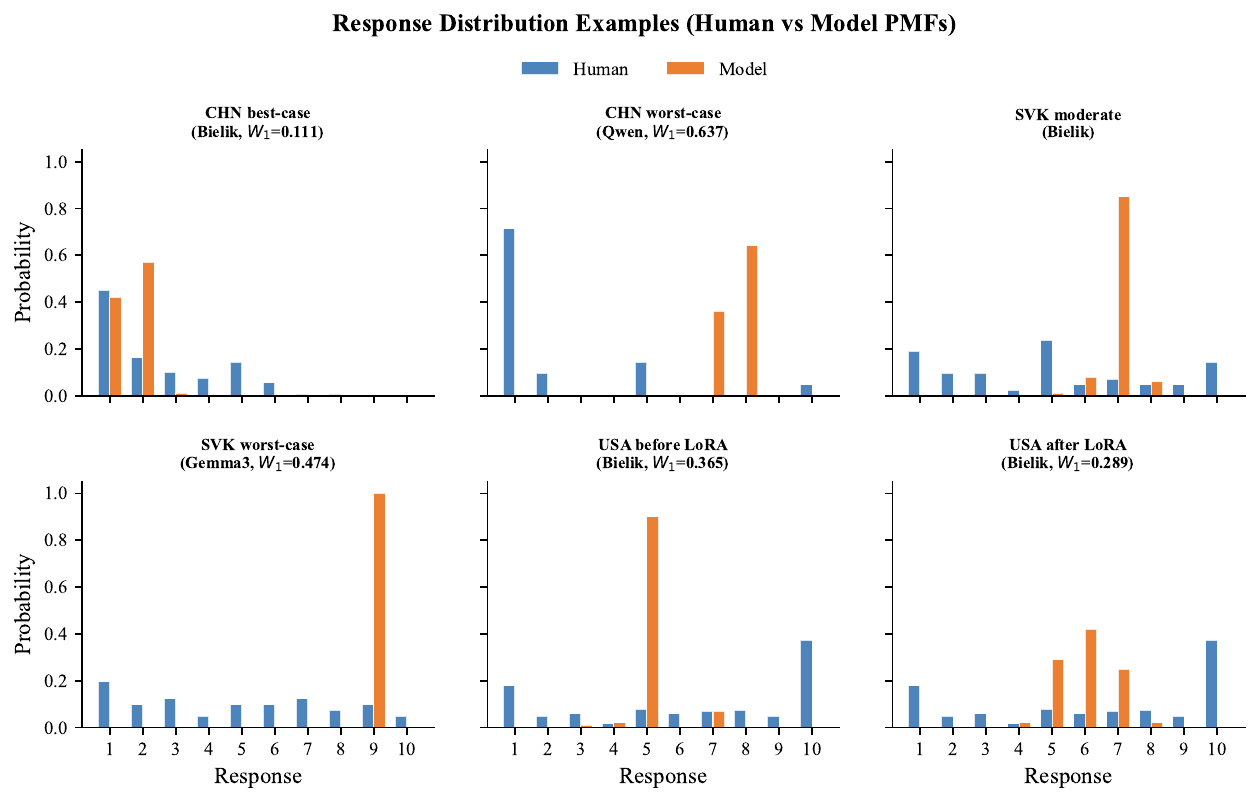}
    \caption{Six example response distribution comparisons (human PMF in blue, model PMF in orange). Cases illustrate: best-case alignment (CHN, Bielik, $\hat{W}_1 = 0.111$), worst-case misalignment (CHN, Qwen, $\hat{W}_1 = 0.637$), moderate alignment (SVK, Bielik), worst Slovak case (SVK, Gemma3, $\hat{W}_1 = 0.474$), and Bielik before vs.\ after LoRA on \texttt{USA\_Male\_30-49\_Medium} ($\hat{W}_1$: $0.365 \to 0.289$).}
    \label{fig:example_pmfs}
\end{figure}

\end{document}